\documentclass[review]{elsarticle}

\usepackage{bm}
\usepackage[version=3]{mhchem} % Package for chemical equation typesetting
\usepackage{siunitx} % Provides the \SI{}{} and \si{} command for typesetting SI units
\usepackage{graphicx} % Required for the inclusion of images
\usepackage{amsmath} % Required for some math elements 
\usepackage[makeroom]{cancel}
\usepackage{amsmath}
\usepackage{algorithm}
\usepackage{booktabs}
\usepackage{amssymb}
\usepackage{url}
\usepackage{rotating}
\usepackage[noend]{algpseudocode}
\usepackage[export]{adjustbox}
\graphicspath{ {Figures/} }
\usepackage{multirow}
\usepackage{pifont}
\usepackage{lineno,hyperref}
\usepackage{subcaption}
\usepackage{xcolor,colortbl}
\usepackage{caption}
\newcommand{\X}{\mathcal{X}}
\newcommand{\Xa}{X^{a_i}}
\newcommand{\x}{\vec{x}}
\newcommand{\ti}{\pmb{\theta}_i}
\newcommand{\tstar}{\pmb{\theta}_*}

\newcommand{\tac}{\theta_{i}^{c}}
\newcommand{\Ua}{U^{a_i}}
\newcommand{\Uac}{U^{(a_i,c)}}

\newcommand{\Da}{D^{a_i}}

\newcommand{\Dc}{D^{y=c}}
\newcommand{\LN}{L_\mathbb{N}}
\newcommand{\dn}{\delta L}
 
\newcommand{\Y}{\mathcal{Y}}
\newcommand{\Ya}{Y^{a_i}}

\newcommand{\R}{\mathcal{R}}
\newcommand{\Us}{U}

\newcommand{\Usage}{\operatorname{usage}}
\newcommand{\Classy}{\textsc{Classy}}
\newcommand{\Cands}{Cands}

\newcommand{\given}{\mid}

\algnewcommand\algorithmicinput{\textbf{Input:}}
\algnewcommand\INPUT{\item[\algorithmicinput]}
\algnewcommand\algorithmicoutput{\textbf{Output:}}
\algnewcommand\OUTPUT{\item[\algorithmicoutput]}
\newcommand{\ra}[1]{\renewcommand{\arraystretch}{#1}}

\DeclareMathOperator*{\argmax}{arg\,max}
\newcommand*{\defeq}{\mathrel{\vcenter{\baselineskip0.5ex \lineskiplimit0pt
			\hbox{\scriptsize.}\hbox{\scriptsize.}}}%
	=}

\renewcommand{\vec}[1]{\mathbf{#1}}

\newcommand{\cmark}{\ding{52}}%
\definecolor{Silver}{rgb}{0.85,0.85,0.85}
\definecolor{Gray}{rgb}{0.5,0.5,0.5}

\newcolumntype{b}{>{\columncolor{Silver}}r}
\makeatletter
\def\ps@pprintTitle{%
	\let\@oddhead\@empty
	\let\@evenhead\@empty
	\def\@oddfoot{}%
	\let\@evenfoot\@oddfoot}
\makeatother

\usepackage{lineno}

\begin{document}

\begin{frontmatter}

\author{Interpretable multiclass classification by MDL-based rule lists}
\address{}

\author[mymainaddress]{Hugo M. Proen\c{c}a\corref{mycorrespondingauthor}}
\cortext[mycorrespondingauthor]{Corresponding author}
\ead{h.manuel.proenca@liacs.leidenuniv.nl}

\author[mymainaddress]{Matthijs van Leeuwen}
\ead{m.van.leeuwen@liacs.leidenuniv.nl}

\address[mymainaddress]{LIACS, Leiden University, The Netherlands}

\begin{abstract}
Interpretable classifiers have recently witnessed an increase in attention from the data mining community because they are inherently easier to understand and explain than their more complex counterparts. Examples of interpretable classification models include decision trees, rule sets, and rule lists. Learning such models often involves optimizing hyperparameters, which typically requires substantial amounts of data and may result in relatively large models.

In this paper, we consider the problem of learning compact yet accurate probabilistic rule lists for multiclass classification. Specifically, we propose a novel formalization based on probabilistic rule lists and the minimum description length (MDL) principle. This results in virtually parameter-free model selection that naturally allows to trade-off model complexity with goodness of fit, by which overfitting and the need for hyperparameter tuning are effectively avoided. Finally, we introduce the \Classy{} algorithm, which greedily finds rule lists according to the proposed criterion.

We empirically demonstrate that \Classy{} selects small probabilistic rule lists that outperform state-of-the-art classifiers when it comes to the combination of predictive performance and interpretability. We show that \Classy{} is insensitive to its only parameter, i.e., the candidate set, and that compression on the training set correlates with classification performance, validating our MDL-based selection criterion.
 
\end{abstract}


\end{frontmatter}

%\linenumbers

\section{Introduction}
\emph{Interpretable machine learning} has recently witnessed a strong increase in attention \cite{doshi17interpretable}, both within and outside the scientific community, driven by the increased use of machine learning in industry and society. This is especially true for applications domains where decision making is crucial and requires transparency, such as in health care \cite{letham2015interpretable,lakkaraju2017learning} and societal problems \cite{lakkarajulearning,zeng2017interpretable}. 

While it is of interest to investigate how existing `black-box' machine learning models can be made transparent \cite{ribeiro2016should}, the trend towards interpretability also offers opportunities for data mining, or \emph{Knowledge Discovery from Data} (KDD), as this field traditionally has a stronger emphasis on intelligibility. 

In recent years several interpretable approaches have been proposed for supervised learning tasks, such as classification and regression. Those include approaches based on prototype vector machines \cite{polaka2017constructing}, generalized additive models \cite{lou2012intelligible}, decisions sets \cite{lakkaraju2016interpretable,wang2016bayesian}, and rule lists \cite{letham2015interpretable,yang2017scalable}.
Restricting our focus to classification, we make two important observations. First, we observe that state-of-the-art algorithms \cite{lakkaraju2016interpretable,wang2016bayesian,letham2015interpretable,yang2017scalable,angelino2017learning} are designed for binary classification; no interpretable methods specifically aimed at multiclass classification have been proposed, in spite of being a common scenario in practice. Multiclass classification is more challenging because of 1)  the increased complexity in model search, due to the uncertain consequences of favouring one class over the others, and 2) the lack of possibilities to prune the search such as commonly used when finding, e.g., decision lists \cite{angelino2017learning} or Bayesian rule lists \cite{yang2017scalable} for binary classification. Our second observation is that although recent methods based on rules \cite{letham2015interpretable,yang2017scalable} and decision sets \cite{lakkaraju2016interpretable,wang2016bayesian} have been shown to be effective, they tend to have 1) a fair number of hyperparameters that need to be fine-tuned, and 2) limited scalability. Especially the need for hyperparameter tuning can be problematic in practice, as it requires significant amounts of computation power and data (i.e., not all data can be used for training, as a substantial part has to be reserved for validation).

To address these shortcomings, \emph{we introduce a novel approach to finding interpretable, probabilistic multiclass classifiers that requires very few hyperparameters and results in compact yet accurate classifiers}. In particular, we will show that our method naturally provides a desirable trade-off between model complexity and classification performance without the need for parameter tuning, which makes the application of our approach very straightforward and the resulting models both adequate classifiers and easy to interpret.

We will use probabilistic rule lists, as both the antecedent of a rule (i.e., a \emph{pattern}) and its consequent (i.e., a probability distribution) are interpretable \cite{letham2015interpretable}. Using a probabilistic model has the additional advantage that one cannot only provide a crisp prediction, but also make a statement about the (un)certainty of that prediction.

We show that, given a set of ordered patterns, we can trivially estimate the corresponding consequent probability distributions from the data. The remaining question, then, is how to select a set of patterns that together define a probabilistic rule list that is accurate yet does not overfit. This is not only important to ensure generalizability beyond the observed data, but also to keep the models as compact as possible: larger models are harder to interpret by a human analyst \cite{huysmans2011empirical}. Recent optimization \cite{lakkaraju2016interpretable} and Bayesian \cite{yang2017scalable} approaches heavily rely on hyperparameters to achieve this, but those need to be tuned by the analyst and we specifically aim to avoid this.

The solution that we propose is based on the minimum description length (MDL) principle \cite{rissanen78,grunwald2007minimum}, which has been successfully used to select small sets of patterns that summarize the data in the context of exploratory data mining \cite{vreeken2011krimp,van2014mining,budhathoki2015difference}. The MDL principle can be paraphrased as ``\emph{induction by compression}'' and roughly states that the best model is the one that best compresses the data. Advantages of the MDL principle include that it has solid theoretical foundations, avoids the need for hyperparameters, and automatically protects against overfitting by balancing model complexity with goodness of fit.

\emph{Our first main contribution is the formalization of the problem of selecting the optimal probabilistic rule list using the minimum description length principle.} Although the MDL principle has been used for pattern-based classification before \cite{vreeken2011krimp}, we are the first to introduce a MDL-based problem formulation aimed at selecting rule lists for multiclass classification. Technically, our approach includes the use of the prequential plug-in code, a form of \emph{refined MDL} that has only been used once in pattern-based modelling \cite{budhathoki2015difference}. One advantage of our approach is that the resulting problem formulation is completely parameter-free.

\emph{Our second main contribution is \Classy, a heuristic algorithm for finding good probabilistic rule lists.} Inspired by the \textsc{Krimp} algorithm \cite{vreeken2011krimp}, we select a good set of rules from a set of candidate patterns. We empirically demonstrate, by means of a variety of experiments, that \Classy{} outperforms RIPPER, C5.0, CART, and Scalable Bayesian Rule Lists (SBRL) \cite{yang2017scalable} when it comes to the combination of classification performance and interpretability, in particular when taking into account that it has much fewer hyperparameters.

\begin{figure}[bth]\centering																
	\ra{1.1}\small \begin{tabular}{@{}llrclr@{}}																		
		\textbf{Rule}	&$		$&	\textbf{antecedent}				&			&	\textbf{consequent}		&	\textbf{usage}		\\ 	\midrule
		1	&$	\textsc{If }	$&$	 \{backbone = no \} 				$&$	 \textsc{ then }		$&$	 \Pr(invertebr.) = 0.55		$&$	10	$	\\ 	
		&$		$&					&			&$	  \Pr(bug) = 0.45 		$&$	8	$	\\ 	
		2	&$	\textsc{else if}	$&$	 \{breathes = no \}				$&$	 \textsc{ then }		$&$	\Pr(fish) = 0.93		$&$	13	$	\\ 	
		&$		$&					&			&$	\Pr(reptile) = 0.07 		$&$	1	$	\\ 	
		3	&$	\textsc{else if}	$&$	\{feathers = yes \} 				$&$	 \textsc{ then }		$&$	 \Pr(bird) = 1.00 		$&$	20	$	\\ 	
		4	&$	\textsc{else if}	$&$	 \{milk = no \} 				$&$	 \textsc{ then }		$&$	\Pr(reptile) = 0.50		$&$	4	$	\\ 	
		&$		$&					&			&$	 \Pr(amphibian) = 0.50 		$&$	4	$	\\ 	
		$\varnothing$	&$	\textsc{else}	$&$	\textsc{else}				$&$	 \textsc{ then }		$&$	 \Pr(mammal) = 1.00 		$&$	41	$	\\ 	
		
	\end{tabular}																		
	\caption{Example of a Probabilistic Rule List (PRL) obtained by \Classy{} on the \emph{zoo} dataset, without the need for any parameter tuning. Test accuracy: $87\%$. The dataset contains $7$ classes, $101$ examples, and $35$ binary variables. Usage refers to the number of examples covered by a certain rule and class label. Note that we did not apply Laplace smoothing here for clarity of presentation; Equation~\eqref{eq:probability} defines the actual probability estimates. }\label{fig:zoo_example}																		
\end{figure}																

To illustrate this, Figure~\ref{fig:zoo_example} shows an example rule list that was found using \Classy{} on the \emph{zoo}\footnote{\url{http://archive.ics.uci.edu/ml/datasets/Zoo}} dataset, without any parameter tuning. Although it is not perfectly accurate, its accuracy ($87\%$) is pretty good considering that there are seven classes and the list only has four rules. Moreover, it provides probabilistic predictions and is very easy to interpret.

The remainder of this paper is organized as follows. First, Section~\ref{sec:relwork} discusses related work, after which we introduce probabilistic rule lists and MDL for such lists in Sections~\ref{sec:rulelists} and \ref{sec:MDL}, respectively. Section~\ref{sec:algorithm} presents the \Classy{} algorithm. After that we continue with experiments in Section~\ref{sec:experiments} and conclude in Section~\ref{sec:conclusion}.

\section{Related work}
\label{sec:relwork}

We start by comparing the most important features of our algorithm to those of state-of-the-art algorithms, and then provide a brief overview of the most relevant literature, grouped into three topics: 1) rule-based models; 2) similar approaches in pattern mining; and 3) MDL-based data mining. For an in-depth overview of interpretable machine learning, we refer to Molnar \cite{molnar2018interpretable}.

Table~\ref{table:contributions} compares the most important features of our proposed approach, called \Classy{}, to those of other rule-based classifiers, which will be described in the next subsections. Classical methods, such as CART \cite{breiman1984classification}, C4.5 \cite{quinlan2014c4} and RIPPER \cite{cohen1995fast}, lack a global optimisation criterion and thus rely on heuristics and hyperparameters to deal with overfitting. Fuzzy rule-based models \cite{alcala2011fuzzy,jimenez2014multi}, here represented by FURIA \cite{huhn2009furia} use rule sets instead of rule lists and lack probabilistic predictions. Recent Bayesian methods \cite{wang2016bayesian,lakkaraju2016interpretable,wang2016bayesian} are limited to small numbers of candidate rules and binary classification, limiting their usability, and are here represented by SBRL \cite{yang2017scalable} (which is representative for all of them). A recent approach also using MDL and probabilistic rule lists (MRL) \cite{aogafinding} is aimed at describing rather than classifying and cannot deal with multiclass problems or a large number of candidates. Interpretable decision sets (IDS) \cite{lakkaraju2016interpretable} and certifiable optimal rules (CORELS) \cite{angelino2017learning} use similar rules but do not provide probabilistic models or predictions.

\bigskip

Note that methods that explain black-box models \cite{ribeiro2016should,ribeiro2018anchors}, typically denoted by the term \emph{explainable machine learning}, also aim to make the decisions of classifiers interpretable. However, they mostly focus on sample-wise (local interpretation) explanations, while we focus on explaining the whole dataset (global interpretation) by means of a single model. As these goals lead to clearly different problem formulations and thus different results, it would not be meaningful to empirically compare our approach to explainable machine learning methods.

\begin{table}[!t]														
	\centering													
	\ra{1.1} \begin{tabular}{@{}lccccc@{}}\toprule													
		Method	&	Multiclass	&	Probabilistic	&	Criterion	&	$\gg$$1$K cand	&	No tuning	\\ 	\midrule
		\Classy	&	\cmark	&	\cmark	&	\cmark	&	\cmark	&	\cmark	\\ 	
		IDS\cite{lakkaraju2016interpretable}	&	\cmark	&	-	&	\cmark	&	\cmark	&	-	\\ 	
		{\small CORELS} \cite{angelino2017learning}	&	-	&	-	&	\cmark	&	-	&	-	\\ 	
		MRL\cite{aogafinding} 	&	-	&	\cmark	&	\cmark	&	-	&	\cmark	\\ 	
		SBRL\cite{yang2017scalable}	&	-	&	\cmark	&	\cmark	&	-	&	-	\\ 	
		FURIA\cite{huhn2009furia}	&	\cmark	&	-	&	-	&	\cmark	&	-	\\ 	
		Others	&	\cmark	&	\cmark	&	-	&	\cmark	&	-	\\ 	\bottomrule
	\end{tabular}													
	\caption{Our approach, \Classy{}, does \emph{Multiclass} classification, makes \emph{Probabilistic} predictions, has a global optimisation \emph{Criterion}, can handle large numbers of \emph{cand}idate rules, and does not need hyperparameter \emph{tuning}. ``Others'' denotes classical algorithms such as CART, CBA, C4.5, and RIPPER. }\label{table:contributions}													
\end{table}

\subsection{Rule-based models}
Rule lists have long been successfully applied for classification; RIPPER is one of the best known algorithms \cite{cohen1995fast}. Similarly, decision trees, which can easily be transformed to rule lists, have been used extensively; CART \cite{breiman1984classification} and C4.5 \cite{quinlan2014c4} are probably the most best-known representatives. These early approaches represent highly greedy algorithms that use heuristic methods and pruning to find the `best' models. 

Fuzzy rules have been extensively studied in the context of classification and interpretability. Several approaches to construct fuzzy rule-based models have been proposed, such as transforming the resulting model of another algorithm into a fuzzy model and posteriorly optimizing it \cite{huhn2009furia}, using genetic algorithms to combine pre-mined fuzzy association rules \cite{alcala2011fuzzy}, and doing a multiobjective search over accuracy and comprehensibility to find Pareto-optimal solutions \cite{jimenez2014multi}. Although these approaches are related, the rules are aggregated in a rule set, i.e., a set of independent if rules that can be activated at the same time to classify one instance, contrary to one rule at the time for rules lists. This makes the comparison between both types of models difficult. Also, these fuzzy rule-based models do not provide probabilistic predictions.

Over the past years, rule learning methods that go beyond greedy approaches have been developed, i.e., by means of probabilistic logic programming for independent rule-like models \cite{bellodi2015structure}, greedy optimization of submodular problem formulation or simulated annealing in the case of decision sets \cite{lakkaraju2016interpretable,wang2016bayesian}, by Monte-Carlo search for Bayesian rule lists \cite{letham2015interpretable,yang2017scalable}, and through branch-and-bound with tight bounds for decision lists \cite{angelino2017learning}.
Even though in theory these approaches could be easily extended to the multiclass scenario, in practice their algorithms do not scale with the higher dimensionality that arises from the search in multiclass space with an optimality criteria. Also, only Bayesian rule lists \cite{letham2015interpretable,yang2017scalable} and Bayesian decision sets \cite{wang2016bayesian} provide probabilistic predictions. 

All previously mentioned algorithms share some similarities with \Classy. In particular Bayesian rule lists \cite{letham2015interpretable,yang2017scalable} are closely related as they use the same type of models, albeit with a different formulation, based on Bayesian statistics. This difference leads to different types of priors---for example, we use the universal code of integers \cite{rissanen1983universal}---and therefore to different results; we will empirically compare the two approaches. Certifiable optimal rules \cite{angelino2017learning} have a similar rule structure but do not provide probabilistic models or predictions. Decision sets \cite{lakkaraju2016interpretable} share the use of rules, but as opposed to (ordered) lists they consider (unordered) sets of rules.

\subsection{Pattern mining}
Association rule mining \cite{agrawal1993mining}, a form of pattern mining, is concerned with mining relationships between itemsets and a target item, e.g., a class. One of its key problems is that it suffers from the infamous \emph{pattern explosion}, i.e., it tends to give enormous amounts of rules. 
Several classifiers based on association rule mining have been proposed. Best-known are probably CBA \cite{ma1998integrating} and CMAR \cite{li2001cmar}, but they tend to lack interpretability because they use large numbers of rules. Ensembles of association rules, such as Harmony \cite{wang2005harmony} or classifiers based on emergent patterns \cite{garcia2014survey}, can increase classification performance when compared to the previous methods, however they can only offer local interpretations. 
Subgroup discovery and similar approaches \cite{novak09sd} are all relevant too, but they focus on finding descriptive patterns and not on finding global classification models. Approaches to supervised pattern mining based on significance testing \cite{webb07sigpatts} do not focus on finding global models either.

A last class of related methods is that of \emph{supervised pattern set mining} \cite{zimmermann14supervised}. The key difference is that these methods do not automatically trade-off model complexity and classification accuracy, requiring the analyst to choose the number of patterns $k$ in advance. 

\subsection{MDL-based data mining}
In data mining, the MDL principle has been used to summarize different types of data, e.g., transaction data \cite{vreeken2011krimp,budhathoki2015difference}, and two-view data \cite{van2015association}.

In prediction it has been previously used to deal with overfitting \cite{cohen1995fast,quinlan2014c4} and in the selection of the best compressing pattern \cite{zhang2000information}.

RIPPER and C4.5 \cite{cohen1995fast,quinlan2014c4} use the MDL principles in their post-processing phase as a criteria for pruning, while we use it in a holistic way for model selection.  
Although Krimp has been used for classification \cite{vreeken2011krimp}, it was not designed for this: it outputs large pattern sets, one for each class, and does not give probabilistic predictions. 
DiffNorm \cite{budhathoki2015difference} creates models for combinations of classes and also uses the prequential plug-in code, but was designed for data summarization. Aoga et al.\ recently also proposed to use probabilistic rule lists and MDL \cite{aogafinding}, but 1) we propose a vastly improved encoding, which is tailored towards prediction (instead of summarization), 2) our solution does multiclass classification, and 3) our algorithm has better scalability.

\section{Multiclass classification with rule lists}
\label{sec:rulelists}

In this section we formalize the probabilistic rule list model and show how to estimate its parameters, i.e., the rule consequent probabilities. The notation most commonly used throughout this paper is summarized in Table~\ref{table:notation}.

Let $D = (X, Y) = \{(\x_1,y_1),(\x_2,y_2),...,(\x_n,y_n)\}$ be a \emph{supervised Boolean dataset}, i.e., a Boolean dataset $X$ with a (multi)class label vector $Y$. Note that any categorical dataset can be trivially represented as a Boolean dataset using dummy variables, hence our methods apply to any categorical dataset as well. Each example forms a pair $(\x,y)$, which consists of an instance of Boolean variables $\x$ and a class label $y$. 

An instance $\x = (x_{1},x_{2},...,x_{k})$ consists of $k=|V|$ binary values, with $V$ the set of all Boolean variables in $X$. That is, $\x$ is an element of the set of all possible Boolean vectors of size $|V|$, i.e, $\x \in \X \doteq  \{0,1\}^{|V|}$. A pattern $a$ is a logical conjunction of variable-value assignments over instance space $\X$, e.g., $a= [x_{2} =1 \wedge x_{3}=1 ]$. A pattern $a$ \emph{occurs} in instance $\x$, denoted $a \sqsubseteq \x $, iff $\x$ satisfies the predicate defined by pattern $a$. Thus, in our example, the pattern occurs in an instance iff $x_{2} =1$ and $x_{3}=1$; the values of the other variables do not influence the occurrence of this pattern. A pattern is said to have size $|a|$ equal to the number of conditions it contains; in this case $|a| = 2$. Each class label is an element of the set of all classes, i.e., $y_i \in \Y$, where $\Y = \{1,\ldots,|\Y|\}$ and $|\Y|$ is the number of classes in the dataset.
 
We consider the problem of multiclass classification. That is, given training data $D$, the goal is to induce a classification model that accurately predicts class label $c \in \Y$ for any (possibly unseen) instance $\x \in \X$.

\subsection{Probabilistic rule lists}

A \emph{probabilistic rule list} (PRL) $R$ is an ordered list of $k$ \emph{rules} $r_1,\ldots,r_k$ ending with a \emph{default rule} $r_\varnothing$, where each defines a probability distribution over the class labels. Note that this means that $R$ has $|R|+1$ rules in total. Each rule consists of a pair $r_i = (a_i,\pmb{\theta} (a_i) )$, where a pattern $a_i$ is the antecedent and a categorical distribution (i.e., a generalized Bernoulli distribution) over the class labels $\pmb{\theta} (a_i)$ is the consequent. Whenever clear from the context we use $\ti$ as shortcut for the parameters associated with pattern $a_i$. Each categorical distribution is parameterized by individual class probabilities $\ti = (\theta_{i}^{c_1},\ldots,\theta_{i}^{c_{|\Y|}})$, such that $\theta_{i}^{c_j} >0, \forall_{i,j}$ and $\sum_j \theta_{i}^{c_j} = 1, \forall_i$. That is, rule $i$ is given by
\begin{equation}\label{eq:categorical}
a_i \rightarrow  y \sim Categorical(\ti).
\end{equation}

The default rule $r_\varnothing$, which intuitively corresponds to a rule with the empty set as antecedent, is associated with a categorical distribution over the class labels denoted by $\pmb{\theta}_\varnothing$. An example PRL with $|R|=2$ rules is given by:
\begin{equation}\label{eq:rulelist}
\begin{split}
\text{rule 1 }: &\textsc{ if }   a_1 \sqsubseteq  \x   \textsc{ then }  y \sim Categorical(\pmb{\theta}_1)\\ 
\text{rule 2 }:& \textsc{ else if }  a_2 \sqsubseteq  \x    \textsc{ then } y \sim Categorical(\pmb{\theta}_2)\\ 
\text{default}:& \textsc{ else }   y \sim Categorical(\pmb{\theta}_\varnothing)\\
\end{split}
\end{equation}

Given a PRL $R$, an instance $\x$ is classified by going through the rule list top-down, i.e., $\x$ is classified according to $r_* = (a_*,\tstar)$, which is defined as the first rule in the list for which $a_*$ occurs in $\x$ ($a_* \sqsubseteq  \x$). If none of the patterns $a_i, \forall_{i \in {1,...,|R|}}$ occurs in a given instance, the classifier automatically falls back to the default rule $r_\varnothing$. From the pattern $a_*$ that is activated for a given instance, the user obtains a corresponding probability distribution $\tstar$ over the class labels for instance $\x$. In case a crisp prediction is necessary, as for example when comparing a PRL with other classifiers, we follow the typical approach, which is to predict the class label that has the highest probability: 
\begin{equation}
\hat{y}= \argmax_{c \in \Y} \theta_{*}^{c}. 
\end{equation}

Note that contrary to decision lists, which only provide an associated class per rule, probabilistic rule lists provide a distribution over all class labels. This provides the user with extra information about the classification that is made, in the form of a probability of seeing each class label for a certain instance. This is especially relevant in the multiclass scenario where crisp classification implies a choice between more than two classes.

\subsection{Parameter estimation} \label{subsec:parest}
%\todo{is ``previous'' correct or should it be ``previous section''?} --> is okay
In the previous section the PRL is assumed to be given, while in practice we want to learn its parameters from the data. We defer the problem of selecting the patterns $a_i$ to the next section and first describe how to estimate the parameters of the categorical distributions from data, i.e., how to estimate $\ti$, for $i \in \{1,\ldots,|R|,\varnothing\}$, given an (ordered) set of patterns.

We first introduce some notation. The \emph{support} of a pattern $a_i$ is the number of times that the pattern occurs in (training) data $D$:
\begin{equation}
supp(a_i) =  \left |  \{\x \subset D \given a_i \sqsubseteq \x \} \right |
\end{equation}  

The \emph{usage} of $a_i \in R$ is the number of times it is activated in (training) data $D$. That is, it is the support of $a_i$ minus the instances that were already covered by other patterns that come before $a_i$ in $R$:
\begin{equation*}\label{eq:usage}
\Usage(a_i \given R,D) = |\{\vec{x} \subset D \given a_i \sqsubseteq \vec{x} \wedge \left ( \bigwedge_{\forall_{j<i}}  a_j \not\sqsubseteq \vec{x}   \right ) 
\}|
\end{equation*}
For ease of presentation, we abbreviate $\Usage(a_i \given R, D)$ as $\Ua$ whenever $D$ and $R$ are clear from the context.

Next, we introduce class-specific usage as the number of times a pattern is activated on a training instance with class label $c$. We define a \emph{class-conditioned dataset} as
\[
\Dc = \{(\x,y) \subset D \given y=c \},
\]
and \emph{class-specific usage} as
\[
\Uac = \Usage(a_i \given R, D^{y=c}).
\]

Given the usages and class-specific usages, which are easy to compute, it is straightforward to define a maximum likelihood estimator for $\Pr(y=c \given a_i)$, for any rule $a_i$ and class $c$. We use a variant that is called a smoothed maximum likelihood estimator:

\begin{equation}\label{eq:probability}
\hat{\theta}_{i}^{c} = \frac{\Uac+\epsilon}{\Ua+ |\Y| \epsilon }.
\end{equation}

Unlike the regular maximum likelihood estimator, this smoothed variant---known as Laplace smoothing---adds a (small) pseudocount $\epsilon$ to each class-specific usage even when that class has no counts. This avoids zero probabilities for any class label and corresponds to using a symmetric Dirichlet prior $\epsilon$ for each class \cite{gelman2013bayesian}. Note that this estimate could be interpreted as the confidence of the $n^{\text{th}}$ rule in a list, accounting for what has been covered by previous rules.

\section{MDL for multiclass classification}
\label{sec:MDL}

Having defined our models and parameter estimator, the remaining question is how to select adequate models. As we are interested in finding compact yet accurate rule lists that do not overfit, we resort to the minimum description length (MDL) \cite{rissanen78,grunwald2007minimum} principle, which can be paraphrased as ``\emph{induction through compression}''. The problem of selecting a concrete rule list from a large space of possible rule lists is a \emph{point hypothesis selection} problem, for which we should use a two-part code \cite{grunwald2007minimum}. 

In contrast to existing pattern-based modeling approaches (e.g., \cite{vreeken2011krimp,van2014mining}), we deal with a \emph{supervised} setting in which the goal is to learn a mapping from instances to class labels. This implies that we are not looking for structure \emph{within} instance data $X$, but for structure in $X$ that helps to \emph{predict} $Y$. 

That is, to induce a mapping from instances to class labels, we should consider the instance data $X$ to be given as `input' to the (classification) model and \emph{only encode the class labels $Y$}. Clearly, the models that we consider are the probabilistic rule lists that we introduced in the previous section. Then, given the complete space of models $\R$, uniquely specified by all ordered sets of patterns over $\X$, the optimal model is the model $R \in \R$ that minimizes

\begin{equation}\label{eq:LengthTotal}
L(D,R) =  L(Y \given X,R)  + L(R),
\end{equation}
where $L(Y \given X,R)$ is the encoded length, in bits\footnote{To obtain lengths in bits, all logarithms in this paper are to the base 2.}, of the class labels given data $X$ and model $R$, and $L(R)$ is the encoded length, in bits, of the model. Equation \eqref{eq:LengthTotal} represents a trade-off between how well the model fits the data, $L(Y \given X,R) $ and the complexity of that model, $L(R)$. Note that, on a high level, two-part code in \eqref{eq:LengthTotal} is similar to the one that was recently used in the context of two-view data \cite{van2015association}, but there the goal was summarization rather than classification and the details of the encodings are very different.  The next subsections describe the two parts of the encoding, together with examples of how to compute the length of the encodings in practice. 

\subsection{Model encoding}

Following the rule of parsimony associated with the MDL principle \cite{grunwald2007minimum}, the model encoding should result in larger code lengths for more complex models. To accomplish this we use only two types of codes for the different model components, the universal code for integers and the uniform code.

The universal code for integers \cite{rissanen1983universal}, also called the universal prior for integers, is given by $\LN(i)= \log k_0 + \log^{\ast} i $, where $\log^{\ast} i = \log i + \log \log i + \ldots$ and $ k_0 \approx 2.865064 $. This code makes no \textit{a priori} assumption about the maximum number $i$ accepted by the model and a small assumption in terms of penalizing larger numbers, as it grows logarithmically with $i$ and thus slower than the number of data instances $n$. This makes it quite different from the Poisson prior typically used in Bayesian approaches \cite{yang2017scalable}: that prior more strongly penalizes integers that are further away from the expectation of the distribution, as defined by the user-chosen parameter. We use $L_\mathbb{N}(n)$ when we want to penalize the increase of elements in the model, such as the number of rules or the length of a pattern. 

The uniform code avoids any bias by assigning code words of equal length to all  elements and is therefore used when all elements are equal. E.g., to encode a variable $x$ from a set of $|V|$ variables: $L_U(x) = - \log \frac{1}{|V|} =\log |V|$.

We will now show how to compute the total length of a model, i.e., a probabilistic rule list $R$ over the variable space $V$: 
\begin{equation} \label{eq:LModel}
L(R) = L_\mathbb{N}(|R|) + \sum_{a_i \subset R} L(a_i),
\end{equation}
where first the number of rules is encoded using the universal code for integers, and then the individual patterns are encoded. The length of pattern $a_i$ is given by
\begin{equation} \label{eq:LengthPattern}
L(a_i) = L_\mathbb{N}(|a_i|) + |a_i| \log|V|
\end{equation}
where the number of conditions in $a_i$ is encoded with the universal code for integers, and then each of its conditions are encoded with a uniform code over $V$. 
Contrary to what is common in existing MDL-based pattern set mining approaches (e.g., \cite{vreeken2011krimp,budhathoki2015difference}), which are aimed at summarization, we do not use normalized supports for encoding our patterns. That is, previous work typically uses codes based on the support of a pattern of size $1$ (singleton), e.g., $a = [ x_2=1 ]$, and normalizes this support by the sum of all their supports. Although this works well for summarization, in classification higher support does not necessarily imply better predictive power; hence we use the uniform code. In general, without prior knowledge the uniform code represents the best, unbiased choice \cite{grunwald2007minimum}.\\

\emph{Example 1 (part 1 of 2)}: We use the example rule list in Figure~\ref{fig:zoo_example} to show how to compute the length of the model encoding. The model contains $4$ rules plus a default one, $1$ condition per rule, over a dataset with $35$ binary variables. According to Equation~\eqref{eq:LModel} the length of the model encoding is:
\begin{equation*}
L(R)   = L_\mathbb{N}(4) + \sum_{i = 1}^{4} L_\mathbb{N}(1) + \log 35  = 31.12 \: \text{bits}\\
\end{equation*}

Note that for the purpose of model selection we are only concerned with the length of the encoding, not in materialised codes, hence the values should not be rounded to natural numbers.
\subsection{Data encoding}

For the encoding of the data we use the \emph{prequential plug-in code}, because it is asymptotically optimal even without any prior knowledge on the probabilities  \cite{grunwald2007minimum}. Moreover, \emph{the prequential plug-in code directly uses and gives us the smoothed maximum likelihood estimates $\tac$ for $\Pr(y=c \given a_i)$}, as defined in Subsection~\ref{subsec:parest}, which makes it a natural choice.

Intuitively, the idea of the prequential plug-in code is that one starts with a pseudocount $\epsilon$ for each possible element, constructs a code using these pseudocounts, starts encoding/sending/decoding messages one by one, and then \emph{updates the count of each element after sending/receiving each individual message}. 

We apply this idea to encode the class labels, $Y$. Ignoring the rule list for a moment, initially each class label has a pseudocount of $\epsilon$. Hence, when sending the first class label, $y_1$, we effectively use a uniform code, i.e., $-\log \frac{\epsilon}{|\Y|\epsilon}$. After that, however, we increase the count of that class label by one. Normalizing the updated counts results in a new categorical probability distribution---hence a new code: $-\log \frac{\epsilon+1}{|\Y|\epsilon+1}$. This code is the \emph{best possible code given the data seen so far} and is equal to the smoothed maximum likelihood of Eq.~\eqref{eq:probability}. Formally, the plug-in code for encoding the class labels is defined as
\begin{equation}\label{eq:ProbPlugIn}
\Pr_{\text{plug-in}} \nolimits (y_i = c \given Y_{i-1}) \defeq  \frac{ |\{y \in Y_{i-1} \given  y =c \}|+\epsilon}{\sum_{k \in \Y}  |\{y \in Y_{i-1} \given y = k \}|+ \epsilon },
\end{equation}
where $y_i$ represents the $i^{\text{th}}$ class label, $Y_{i-1} = \{y_1,...,y_{i-1}\}$ represents the sequence of the $i-1$ first class labels, and $\epsilon$ is the pseudocount necessary for $\Pr_{\text{plug-in}} (y_1 = c \given  y_0)$ to be valid. Choosing the uniform prior, i.e., $\epsilon = 1$, is a common choice for categorical distributions \cite{yang2017scalable}, and we will use that value henceforth. 

We now show how the probabilistic rule lists can be used in the encoding of the class labels. By definition, only one rule is activated for each instance, hence each rule only activates in a unique part (subset) of the dataset. By realizing that the encoding of an example in a subset only depends on the rule that formed that subset, the encoding of the dataset can be simplified to the sum of the encoding of its subsets. We define the part covered by a rule antecedent $a_i \in R$ as 
\[
\Da =\{\Xa,\Ya  \} = \{(\x,y) \in D \given a_i \sqsubseteq \x \wedge \left ( \bigwedge_{\forall_{j<i}}  a_j \not \sqsubseteq \vec{x}   \right ) \}.
\]

Thus it is possible to define the encoding of the whole data as:

\begin{equation}\label{eq:LData}
L(Y \given D,R) = \sum_{a_i \in R} L(\Ya \given \Xa, R).
\end{equation}

Inserting the prequential plug-in code \eqref{eq:ProbPlugIn} in \eqref{eq:LData} we obtain:
\begin{equation}\label{eq:PartEncoding}
\begin{split}
L(\Ya \given \Xa,R)  & = - \log \left (\prod_{j=1}^{\Ua} \Pr_{\text{plug-in}} (y_j \given \Ya_{j-1}) \right )  \\
& = - \log \left ( \frac{ \prod_{c \in \Y} \prod_{j=0}^{\Uac -1} (j+ \epsilon) }{\prod_{j=0}^{\Uac-1} (j+ \epsilon C) } \right )  \\
& =- \log \left (\frac{ \prod_{c \in \Y} (\Uac -1 + \epsilon)!/( \epsilon-1) !}{(\Ua -1 + \epsilon |\Y|)! / (\epsilon |\Y| -1)! }  \right ) \\
& =- \log \left (\frac{ \prod_{c \in \Y} \Gamma (\Uac + \epsilon) /\Gamma(\epsilon) }{\Gamma (\Ua + \epsilon |\Y|) /{\Gamma ( \epsilon |\Y|) }}  \right ),
\end{split}
\end{equation}
where $\Ya_{j}$ is a sequence of class labels of length $j$ in part $\Da$, and $\Ua$ and $\Uac$ are the usage and class-specific usage of pattern $a_i$ respectively. Further, $\Gamma$ is the gamma function, an extension of the factorial to real and complex numbers that is given by $\Gamma(i) = (i-1)!$. 

Even though the code was formulated for sequential data, the order in which the class labels are transmitted in \textit{i.i.d} data does not affect the encoded length, as the probability distribution only depends on the \emph{usage} of the patterns, not on the order, as can be seen in the last equation. \\

\emph{Example 1 (part 2 of 2)}: We continue our example of Figure~\ref{fig:zoo_example} with the computation of the length of the data encoding. For this we use $\epsilon = 1$, as we will also use in our experiments. First, we show how to compute the length of encoding the part covered by rule $1$ using Equation~\eqref{eq:PartEncoding}: 
\begin{equation*}
L(Y^{a_{1}} \given X^{a_{1}} ,R)  =- \log \left (\frac{\Gamma (10+1)\Gamma (8+1)\Gamma (1)^{5}}{\Gamma(1)^{7}} \cdot \frac{\Gamma (7)}{\Gamma (18 + 7)}  \right ) = 32.46 \: \text{bits}
\end{equation*}

We repeat this step (not shown) for the other parts of the dataset, which are covered by rules $2,3,4$ and $\varnothing$ respectively. Summing the length of all five parts, following Equation~\eqref{eq:LData}, we obtain $110.36$ bits for the length of the data encoding given $R$, i.e., $L(Y \given D,R)$.

Finally, we sum the results obtained for $L(Y \given D,R)$ and $L(R)$ (in the first part of this example), and obtain $ 141.48$ bits for the joint length of the data and model encoding, as defined by Equation~\eqref{eq:LengthTotal}. 

\begin{table}[!h]
	\centering					
	\small \ra{1.1} \begin{tabular}{@{}lll@{}}\toprule					
		Symbol	&	&	Definition	\\	\toprule
		$	D	$&	&	Dataset of examples	\\	
		$	D^{a}	$&	&	Subset of $D$ where pattern $a$ occurs	\\	
		$	D^{y=c}	$&	&	Subset of $D$ where class label $c$ occurs	\\	
		$	D^{(a,y=c)}	$&	&	Subset of $D$ where $a$ and $c$ both occur	\\	
		$	n	$&	&	Total number  of examples/instances	\\	
		$	X	$&	&	Dataset of instances of $D$ (without class labels)	\\	
		$	V	$&	&	Set of all Boolean variables in $D$	\\	
		$	|V|	$&	&	Total number of Boolean variables in $D$	\\	
		$	\mathcal{X}	$&	&	Set of all possible binary  vectors of size $|V|$	\\	
		$	\vec{x}	$&	&	Instance 	\\	
		$	x	$&	&	Boolean variable	\\	
		$	Y	$&	&	Vector of class labels in $D$	\\	
		$	\mathcal{Y}	$&	&	Set of all class labels in $D$	\\	
		$	|\mathcal{Y}|	$&	&	Total number of class labels in $D$	\\	
		$	y	$&	&	Class label 	\\	
		$	\mathcal{R}	$&	&	Set of all proabilistic rule lists	\\	
		$	R	$&	&	Proabilistic rule list	\\	
		$	|R|	$&	&	Number of rules in $R$ (excluding the default rule)	\\	
		$	r_i 	$&	&	$i^{th}$ rule of $R$	\\	
		$	a_i	$&	&	Pattern/rule antecedent of $r_i$	\\	
		$	|a_i|	$&	&	Number of logical conditions in pattern $a_i$	\\	
		$	supp(a)	$&	&	Number of instances where $a$ occurs	\\	
		$	U^{a_i} 	$&	&	Usage of $a_i$ given $R$	\\	
		$	U^{(a_i,c)}	$&	&	Usage of $a_i$  where class $c$ also occurs given $R$	\\	
		$	\ti	$&	&	Vector of probabilities of each class label of $r_i$	\\	
		$	\tac	$&	&	Probability of the $i^{th}$  rule for class $c$	\\	\bottomrule
	\end{tabular}					
	\caption{Table of commonly used notation.}\label{table:notation}					
\end{table}

\section{The \Classy{} algorithm}
\label{sec:algorithm}

Given our model class---probabilistic rule lists---and its corresponding MDL formulation, what remains is to develop an algorithm that---given the training data---finds the best model according to our MDL criterion. To this end, in this sectio, we present \Classy{}, a greedy search based algorithm that iteratively finds the best rules to add to a rule list. 
This section is structured as follows. First, a brief description of separate-and-conquer greedy search is given. Then compression gain, i.e., the measure that uses compression to score candidate rules, is described. After that, the \Classy{} algorithm is defined. Then, it is explained how individual rules---candidates for the model---are generated from the data. Finally, we analyse \Classy{}'s time and space complexity.

\subsection{Separate-and-conquer greedy search}

Greedy search is very commonly used for learning decision trees and rule lists \cite{quinlan2014c4,cohen1995fast,furnkranz2012foundations}, as well as for pattern-based modelling using the MDL principle \cite{vreeken2011krimp,budhathoki2015difference,van2015association}. A few recent approaches use optimization techniques \cite{yang2017scalable}, but these have the limitation that the search space must be strongly reduced, providing an exact solution to an approximate problem (as opposed to an approximate solution to an exact problem). 

Global heuristics, such as evolutionary algorithms, have been extensively applied to fuzzy rule-based model learning \cite{fernandez2015revisiting}, and although they could also be applied here, we found that the arguments in favor of a local search approach were stronger: 1) local heuristics have often been successfully applied for pattern-based modelling using the MDL principle, making it a natural approach to consider; 2) local heuristics are typically faster than global heuristics, as much fewer candidates need to be evaluated; 3) global heuristics typically require substantially more (hyper)parameters that need to be tuned (e.g., population size, selection and mutation operators, etc.), while local heuristics have very few. 

Given the arguments presented here the algorithm that we propose is based on greedy search. More specifically, it is a heuristic algorithm that, starting from a rule list with just a default rule equal to the priors of the class labels in the data, adds rules according to the well-known \emph{separate-and-conquer} strategy \cite{furnkranz2012foundations}: 1) iteratively find and add the rule that gives the largest change in compression; 2) remove the data covered by that rule; and 3) repeat steps 1-2 until compression cannot be improved. This implies that \emph{we always add rules at the end of the list, but before the default rule}. 

\subsection{Compression gain}

The proposed heuristic is based on the compression gain that is obtained by adding a rule $r = (a,\pmb{\theta})$ to a rule list $R$, which will be denoted by $R \oplus r$. We will argue---and demonstrate empirically later---that for the current task it is better to consider \emph{normalized} gain rather than the typically used \emph{absolute} gain. Note that the gains are defined positive if adding a rule represents a compression improvement, and negative vice-versa.

\textbf{Absolute compression gain}, denoted $\Delta L(D,R \oplus r)$, is defined as the difference in code length before and after adding a rule $r$ to $R$. 
The gain can be divided in two parts:  \emph{data gain}, $\Delta L(Y \given X,R \oplus r)$, and \emph{model gain}, $\Delta L(R \oplus r)$. Together this gives
\begin{equation}\label{eq:gain}
\begin{split}
\Delta L(D,R \oplus r)  & =  L(D,R) -L(D,R \oplus r) \\
& = \underbrace{L(Y \given X,R)-L(Y \given X,R \oplus r)}_{\Delta L(Y\given X,R \oplus r)} \\
& +  \underbrace{ L(R)-L(R \oplus r) }_{\Delta L(R)}.
\end{split}
\end{equation} 

Using Eq.~\eqref{eq:LModel} we show absolute gain as: 
\begin{equation}\label{eq:gainmodel}
\begin{split}
\Delta L(R \oplus r)   = &  \LN(|R|) - \LN(|R|+1)   \\
& -\LN(|a|) -|a| \log|V|.
\end{split}
\end{equation}

Note that model gain is always negative, as adding a rule adds additional complexity to the model. 

In the case of the data gain it should be noted that adding rule $r$ to $R$ only activates the part of the data previously covered by the default rule, as new rules are only added after the previous ones and before the default rule. This search strategy of adding rules assumes that the previous rules already cover their subset well, and that improvements only need to be made where no rule is activated, which corresponds to the region of the dataset covered by the default rule. Hence, we only need to compute the difference in length of using the previous default rule $\varnothing$ and the combination of the new pattern $a \in r$ with the new default rule $\varnothing'$. Using Equation~\eqref{eq:LData} we obtain

\begin{equation}\label{eq:deltafinal}
\begin{split}
\Delta L(Y \given X,R \oplus r)  &=\overbrace{\cancel{\sum_{a_i \in R} L(\Ya \given \Xa,R)} +  L(Y^{\varnothing} \given X^{\varnothing},R \oplus r)}^{L(Y \given X,R)}.\\
&  \underbrace{ -\cancel{ \sum_{a_i \in R} L(\Ya \given \Xa,R)}   -L(Y^{\varnothing'} \given X^{\varnothing'},R) - L(Y^{a} \given Y^{a},R \oplus r)}_{L(Y \given X,R \oplus r)} \\ 
\end{split}
\end{equation}

\textbf{Normalized compression gain}, denoted  $ \dn (D,X \oplus r)$, is defined as the absolute gain normalized by the number of instances that are activated by pattern $a \in r$, which can be obtained by dividing absolute gain by the usage of $a$:
\begin{equation}\label{eq:normalizedgain}
\dn (Y \given X,R \oplus r) =  \frac{\Delta L(Y \given X,R \oplus r) }{ \Us^{a}}
\end{equation}
By normalizing for the number of instances that a rule covers, normalized gain \emph{favors rules that cover fewer instances but provide more accurate predictions} compared to absolute gain. When greedily covering the data, it is essential to prevent choosing large but moderately accurate rules in an early stage; this is likely to lead to local optima in the search space, from which it could be hard to escape. As this is bound to happen when using absolute gain, our hypothesis is that normalized gain will lead to better rule lists. We will empirically verify if this is indeed the case.

\subsection{Candidate generation}\label{section:CandidateGeneration}

Candidates are probabilistic rules of the form $r=(a,\pmb{\theta})$ that are considered for addition to a rule list for a dataset $D$. The candidates are generated by first mining a rule antecedent/pattern $a$ using a standard frequent pattern mining algorithm, e.g., FP-growth \cite{borgelt2003efficient}, and then finding the corresponding consequent categorical distribution $\pmb{\theta}$ given the dataset, i.e., using Equation~\eqref{eq:probability}. In practice these mining algorithms have only two parameters: the minimum support threshold $m_s$ and the maximum length $l_{max}$ of a pattern. Mining frequent patterns can be done efficiently due to the anti-monotone property of their support, i.e., given a pattern $a$ and $b$, if $a$ has less conditions then $b$, i.e., $a \subset b$, implies that $supp(a) \geq supp(b)$. This property is also used to \textbf{remove strictly redundant rules} in \Classy{}. Given all candidates from the frequent pattern mining algorithm, if the antecedent $a$ is a strict subset of antecedent $b$, i.e., $a \subset b$, and they have equal support, $supp(a) = supp(b)$, we say that antecedent $b$ is redundant and will never be selected. This is a consequence of their encoding, i.e., $L_{plug-in}(Y^{a} \given X^{a} ,R) = L_{plug-in}(Y^{b} \given X^{b} ,R)$ in the case they are being considered for the same position, and that the model encoding length of $b$ will always be larger than $a$, i.e., $L(a) <L(b)$. From this we can conclude that $b$ will never be preferred over $a$ during model search, as the gain of $a$ will always be greater. 

\subsection{Finding good rule lists}
We are now ready to introduce \Classy{}, a greedy algorithm for finding good solutions to the MDL-based multiclass classification problem as formalized in Section~\ref{sec:MDL}. The algorithm, outlined in  Algorithm~\ref{alg:PDL}, expects as input a (supervised) training dataset $D$ and a set of candidate patterns, e.g., a set of frequent itemsets mined from $D$, and returns a probabilistic rule list.

The first step of our algorithm, line~\ref{alg:red} is to remove the strictly redundant patterns as mentioned in Section~\ref{section:CandidateGeneration}. After that, in line~\ref{alg:initialize} we initialize the rule list with the default rule, which acts as the baseline model to start from. Then, while there is a rule that improves compression (Ln~\ref{alg:loop}), we keep iterating over three steps: 1) we select the best rule to add (Ln~\ref{alg:gain})---we here use normalized gain for ease of presentation, but this can be trivially replaced by absolute gain; 2) we add it to the rule list (Ln~\ref{alg:addrule}); and 3) we update the usage, and gain of the candidate list (Ln~\ref{alg:update}). To update the usage of a candidate it is necessary to remove from its usage the instances that it has in common with the previous added rule, and then the gain of adding the candidate can be updated. When there is no rule that improves compression (negative gain) the while loop stops and the rule list is returned.

\begin{algorithm}
	\caption{The \Classy{} algorithm}\label{alg:PDL}
	\begin{algorithmic}[1]
		\INPUT Dataset $D$, candidate set $\Cands$
		\OUTPUT Multiclass probabilistic rule list $R$
		%		\State $\vec{X},\vec{N} \gets \Cands$ \label{alg:Eclat}
		\State $\Cands \gets RemoveRedundancy(\Cands)$ \label{alg:red}
		\State $R \gets [\varnothing]$ \label{alg:initialize}
		\Repeat \label{alg:while}
			\State $r \gets \argmax_{\forall r' \in \Cands} :\dn (D,R \oplus r')$ \label{alg:gain}
		\State $R  \gets R \oplus r$	\label{alg:addrule}	
		\State 	$\operatorname{Update Candidates}(D, R, \Cands)$ 		\label{alg:update} 
		\Until{$\dn (D,R \oplus r') \leq 0, \forall r' \in \Cands$}\label{alg:loop}
		\State \textbf{return} $R$
	\end{algorithmic}
\end{algorithm}

\subsection{Complexity}\label{section:complexity}

In this section we analyze the time and space complexity of \Classy{}.
In terms of time complexity, \Classy{} can be divided in two parts: 1) an initialization step, and 2) an iterative loop where one rule is added to a PRL in each iteration.

The time complexity of the initialization step is dominated by sorting the candidates (ascending by length) obtained after running the frequent pattern mining algorithm, and the computation of their instance ids, i.e., the indexes of the instances where each candidate is present. Sorting all candidates takes $\mathcal{O}(|Cands|\log|Cands|)$ time. To compute the instance ids of the candidates, \Classy{} first computes the presence of each singleton condition, i.e., $x_i = 1$ is tested for each variable, in each instance, and then stores them as a bitset in a hash table. As this is done for the whole dataset, it takes $\mathcal{O}(|D||V|)$ time. Then, for candidates of size equal or greater than two and given a sorted array of candidates, it sequentially computes the instance ids of each candidate $a$ based on its decomposition in two candidates of one less condition, i.e., it computes the ids of $a$ based on two candidates $b_1$ and $b_2$ for which $b_1 \cup b_2 = a$ of length $|b_1|=|b_2|= |a|-1$. The ids of $a$ are obtained by the intersection of the sets of instance ids of the smaller length candidates and has a complexity of $\mathcal{O}( |(b_1)_{ids}|+|(b_2)_{ids}|)$. As this is done for each class and in a worst case it would cover the whole dataset, it takes $\mathcal{O}(|D|+ |D|)$. Doing this for all candidates gives $\mathcal{O}(|Cands| |D|)$.

After the initialization step \Classy{} iteratively finds the best rule to add for a total of $|R|$ runs, where $R$ is the PRL that \Classy{} outputs at the end. The time complexity of this loop is dominated by the removal of the instance ids that all candidates have in common with the last added rule. Using again the fact that the intersection of instance ids is upper bounded by the dataset size $|D|$, the removal of instance ids takes at most $\mathcal{O}(|R||Cands||D|)$ time. Given that the rule list can grow at most to the size of the dataset, an upper bound on this complexity is $\mathcal{O}(|Cands||D|^2)$.  
Joining everything together, \Classy{} has a worst case time complexity of
\begin{equation*}
\mathcal{O}(|Cands| |D|^2),
\end{equation*}
which is really a worst case scenario, because in general MDL will obtain PRLs that are much smaller than the dataset size, i.e., $|R| \ll |D|$, making it possible to treat it as a constant. Making this assumption we obtain a more realistic worst case time complexity of
\begin{equation*}
\mathcal{O}(|Cands|\log|Cands| +|Cands| |D|).
\end{equation*}

Note that the time complexity associated with the Gamma function used in the computation of lengths \eqref{eq:LData} and gains \eqref{eq:normalizedgain} of data encoding is not problematic when compared with the other terms. This is due to its recursive computation for $|D|$ values, which can be stored in a dictionary. In total this takes $\mathcal{O}(M(|D|) + |D|)$ time, where $M(*)$ is the complexity of the used multiplication; in the case of our Python implementation this is the Katsuraba multiplication. From then on, the lookup of a value only takes $\mathcal{O}(1)$ time.

In terms of memory complexity, \Classy{} has to store for each candidate for each class: their instance ids $\mathcal{O}(|(a)_{ids}||\Y|)$, their support $\mathcal{O}(|\Y|)$, and their score $\mathcal{O}(|\Y|)$. It is easy to see that $|(a)_{ids}||\Y|$ is upper bounded by the dataset size $|D|$ and that all other memory requirements will be dominated by this part. Also, the storage of the gamma function for each integer up to $|D|$ is only $\mathcal{O}(|D|)$, which gets dwarfed by the instance storage, thus obtaining a worst case memory complexity of
\begin{equation*}
\mathcal{O}(|D||Cands|).
\end{equation*}

\section{Experiments}
\label{sec:experiments}

In this section we empirically evaluate our approach\footnote{Implementation available on \url{https://github.com/HMProenca/MDLRuleLists}}, first in terms of its sensitivity to the candidate set provided and the relationship between compression and classification performance, and second in comparison to a set of representative, state-of-the-art baselines in terms of classification performance, interpretability, overfitting, and runtime.

\textbf{Data}. We use $17$ varied datasets (see Table~\ref{table:datasets}) from the LUCS/KDD\footnote{\url{http://cgi.csc.liv.ac.uk/~frans/KDD/Software/LUCS-KDD-DN/DataSets/dataSets.html}} repository, all of which are commonly used in classification papers. They were selected to be diverse, ranging from $150$ to $48\,842$ samples, from $16$ to $157$ Boolean variables, and from $2$ to $18$ classes. 

\textbf{Candidates}. Frequent pattern mining algorithms generate different candidate sets by setting different values for the minimum support per class threshold $m_{s}$ and maximum pattern length $l_{max}$. To demonstrate that \Classy{} is insensitive to the exact settings of these parameters, we fix a single set of parameter values for all experiments on all datasets (except when we investigate the influence of the candidate set). Specifically, we use $l_{max}=4$ and $m_s=5\%$, to obtain a desirable trade-off between candidate set size, convergence, and runtime. 

These values were objectively derived based on two criteria: making each run finish within $10$ minutes while demonstrating that \Classy{} can deal with large candidate set sizes. First we chose $l_{max}=4$ because this potentially results in very large candidate sets with many redundant rules (i.e., rules that are very similar / strongly overlapping). We then fixed $m_s$ by requiring the runs for all datasets to strictly finish in under $10$ minutes and for most datasets even under $1$ minute, so as to be comparable to $CART$, $C5.0$ and $JRip$ in runtime, and also to have attained (empirical) convergence in terms of compression ratio on the training set---further lowering $m_s$ would not increase compression---as can be seen from the vertical dashed lines in Figure~\ref{fig:candidatesetinfluence}c.

Candidate patterns are mined using Borgelt's implementation of the well-known frequent pattern mining algorithm FP-growth \cite{borgelt2003efficient}. The same candidate set was used for all experiments except when assessing its influence on \Classy{} in Section~\ref{section:candidatesetinfluence}. For that experiment we fixed $l_{max}=4$ and varied the minimum support threshold per class from $m_{s} = \{0.1\%,0.5\%,1\%,$ $2\%,5\%,10\%,15\%,20\%,25\% \}$.

\textbf{Evaluation criteria}. We evaluate and compare our approach based on classification performance, overfitting, interpretability, and runtime. In addition, we assess the influence of the candidate set on our algorithm and whether better compression corresponds to better classification. All results presented are averages obtained using $10$ times repeated $10$-fold cross-validation (with different seeds). 

To quantify how well a rule list compresses the class labels, we define \emph{relative compression} as 
\begin{equation}\label{eq:relcompress}
L \% = \frac{L(D,R)}{L(D, \{\varnothing\})},
\end{equation}
where $L(D, \{\varnothing\})$ is the compressed size of the data given the rule list with only a default rule, i.e., with only the dataset priors for each class. We measure relative compression on the training data, as we use that for model selection.

Classification performance is measured using three measures: accuracy: balanced accuracy \cite{brodersen2010balanced}; and \emph{Area Under the ROC Curve} (AUC). Each measure portrays different aspects of the classifier performance. Accuracy shows the total number of correct classifications. Balanced accuracy, or averaged class accuracy, takes into account the imbalance of class distributions in the dataset and gives the same importance to each class. It is obtained by averaging the recall associated with each class label; informally:
\begin{equation*}
bAcc = \frac{1}{|\Y|} \sum_{c \in \Y} \text{recall}(c)
\end{equation*}

AUC, on the other hand, is not based on a fixed threshold and takes into account the probabilities associated with each prediction. In case of multiclass datasets we use \emph{weighted AUC} \cite{provost2000well}, as it takes into account the class distribution in the dataset. Weighted AUC is obtained by weighing per-class `binary' AUCs (one-versus-all) with the marginal class frequencies:
\begin{equation}
AUC_{weighted} = \sum_{c \in \Y} AUC(c) \frac{supp(c)}{|D|},
\end{equation}
where $AUC(c)$ is the one-versus-all AUC for class label $c$ and $\frac{supp(c)}{|D|}$ is the frequency of that same class label.

%Classification performance is measured using the \emph{Area Under the ROC Curve} (AUC), as we want to evaluate how well the probabilistic models separate the classes. For multiclass datasets we use \emph{weighted AUC} \cite{provost2000well}, obtained by weighing per-class binary AUCs (one-versus-all) with the marginal class frequencies:

For interpretability we follow the most commonly used interpretation, i.e., that smaller models are easier to understand \cite{doshi17interpretable}. With this in mind, we assess: the number of rules and the number of conditions per rule; in all cases, fewer is better. When analyzing decision trees, the number of leaves is given as the number of rules (which includes the default rule), and the average depth of the leaves (except for the longest---assumed the default rule) is given as the number of conditions per rule. Although rule lists derived from decision trees can often be simplified, we here choose not to do this because these directly measures how it would be read by humans.

Overfitting is measured in terms of the absolute difference between the AUC performance in the training set and in the test set. Finally, for runtime, wall clock time in minutes is measured; no parallelization was used. 

\textbf{Note on standard deviations}. Given that the number of values reported both in figures and tables is large, and that standard deviations are usually $2+$ orders of magnitude smaller than their corresponding averages, we choose not to report them---to avoid unnecessarily cluttering the presentation. We did analyse them though and explicitly comment on the few cases where relevant.

\begin{table}[!h]\centering  										
	\ra{1.1} \begin{tabular}{@{}lrrrr@{}}\toprule										
		Dataset	&	$|D|$	&	$|V|$	&	$|\Y|$	&	$|\Cands|$		\\  \midrule
		hepatitis 	&$	155	$&$	48	$&$	2	$&$	39137	$	\\ 
		ionosphere 	&$	351	$&$	155	$&$	2	$&$	332560	$	\\ 
		horsecolic 	&$	368	$&$	81	$&$	2	$&$	23552	$	\\ 
		cylBands 	&$	540	$&$	120	$&$	2	$&$	304749	$	\\ 
		breast 	&$	699	$&$	14	$&$	2	$&$	299	$	\\ 
		pima 	&$	768	$&$	34	$&$	2	$&$	543	$	\\ 
		tictactoe 	&$	958	$&$	26	$&$	2	$&$	1907	$	\\ 
		mushroom 	&$	8124	$&$	84	$&$	2	$&$	79602	$	\\ 
		adult 	&$	48842	$&$	96	$&$	2	$&$	7231	$	\\  
		iris 	&$	150	$&$	14	$&$	3	$&$	144	$	\\ 
		wine 	&$	178	$&$	63	$&$	3	$&$	13439	$	\\ 
		waveform 	&$	5000	$&$	96	$&$	3	$&$	86889	$	\\ 
		heart 	&$	303	$&$	46	$&$	5	$&$	21876	$	\\ 
		pageblocks 	&$	5473	$&$	39	$&$	5	$&$	2902	$	\\ 
		led7 	&$	3200	$&$	22	$&$	10	$&$	2507	$	\\ 
		pendigits 	&$	10992	$&$	81	$&$	10	$&$	107001	$	\\ 
		chessbig 	&$	28056	$&$	54	$&$	18	$&$	1384	$	\\ 
		\bottomrule										
	\end{tabular}										
	\caption{Dataset properties: number of \{samples, binary variables, classes, average number of candidate patterns per fold for \Classy{} with $m_s=5\%$ and $l_{max}=4$\}. The datasets are ordered first by number of classes and then by the number of samples.}\label{table:datasets}										
\end{table}

\subsection{Compression versus classification}\label{section:compressiongain}
\begin{figure}[!t]
	\centering
	\includegraphics[clip,width=1\textwidth]{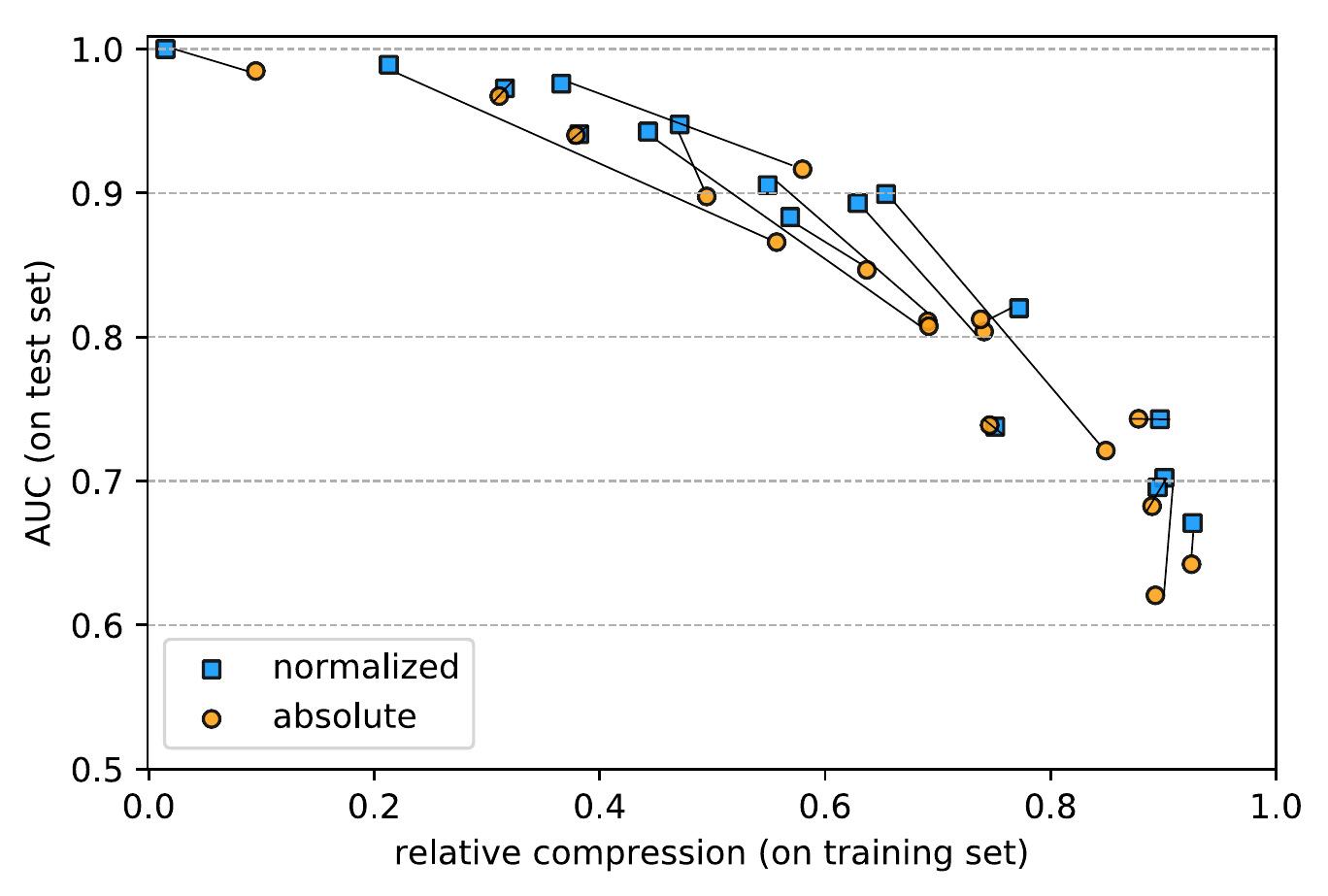}
	\caption{Relation between compression and AUC; better compression on the training set (lower relative compression) corresponds to better classification on the test set (higher AUC). Obtained with \Classy{} using normalized (squares) and absolute (circles) gain, on all 17 datasets;
	each point represents the $10$ times repeated $10$-fold average for one dataset with one type of gain; each connected pair represents the same dataset, for the two types of gain.}
	\label{fig:RelativeCompression}
\end{figure}

We first investigate the effect of using absolute \eqref{eq:gain} or normalized gain \eqref{eq:normalizedgain}. To this end Figure~\ref{fig:RelativeCompression} depicts how the two heuristics perform with respect to relative compression (on the training set) and AUC (on the test set). 

The first observation is that better compression of the training data clearly corresponds to better classification performance on the test data. This is backed by a correlation of $-0.92$ and a corresponding p-value lower than $0.0001$ for the independence test between both variables for the normalized gain data. This is a crucial observation, as it constitutes an independent, empirical validation of using the MDL principle for rule list selection. Moreover, it also shows that MDL successfully protects against overfitting: using normalized gain leads to models that not only compress the training data better, but also provides accurate predictions on the test data.

The second observation is that normalized gain performs better overall than absolute gain: AUC is higher in $15$ out of $17$ cases and relative compression is lower or equal in $11$ out of $17$ times. This confirms that normalized gain is, as we hypothesized, the best choice. We will therefore use \emph{normalized gain} for the remaining experiments.

\subsection{Candidate set influence}\label{section:candidatesetinfluence}
In this set of experiments we study the influence of the candidate set on \Classy{}, which technically is its only ``parameter'', as it is the only part that can influence its output given the same dataset.
In order to vary the candidate set objectively, the minimum support threshold ranges over $m_{s} = \{0.1\%, 0.5\%, 1\%, 2\%, 5\%, 10\%,$ $15\%,  20\%, 25\% \}$ and the maximum pattern length was fixed at $l_{max} = 4$, allowing the generation of large candidate sets. 

\begin{figure*}[!h]
	\centering
	\begin{subfigure}{\textwidth}
		\centering
		\includegraphics[width=\linewidth]{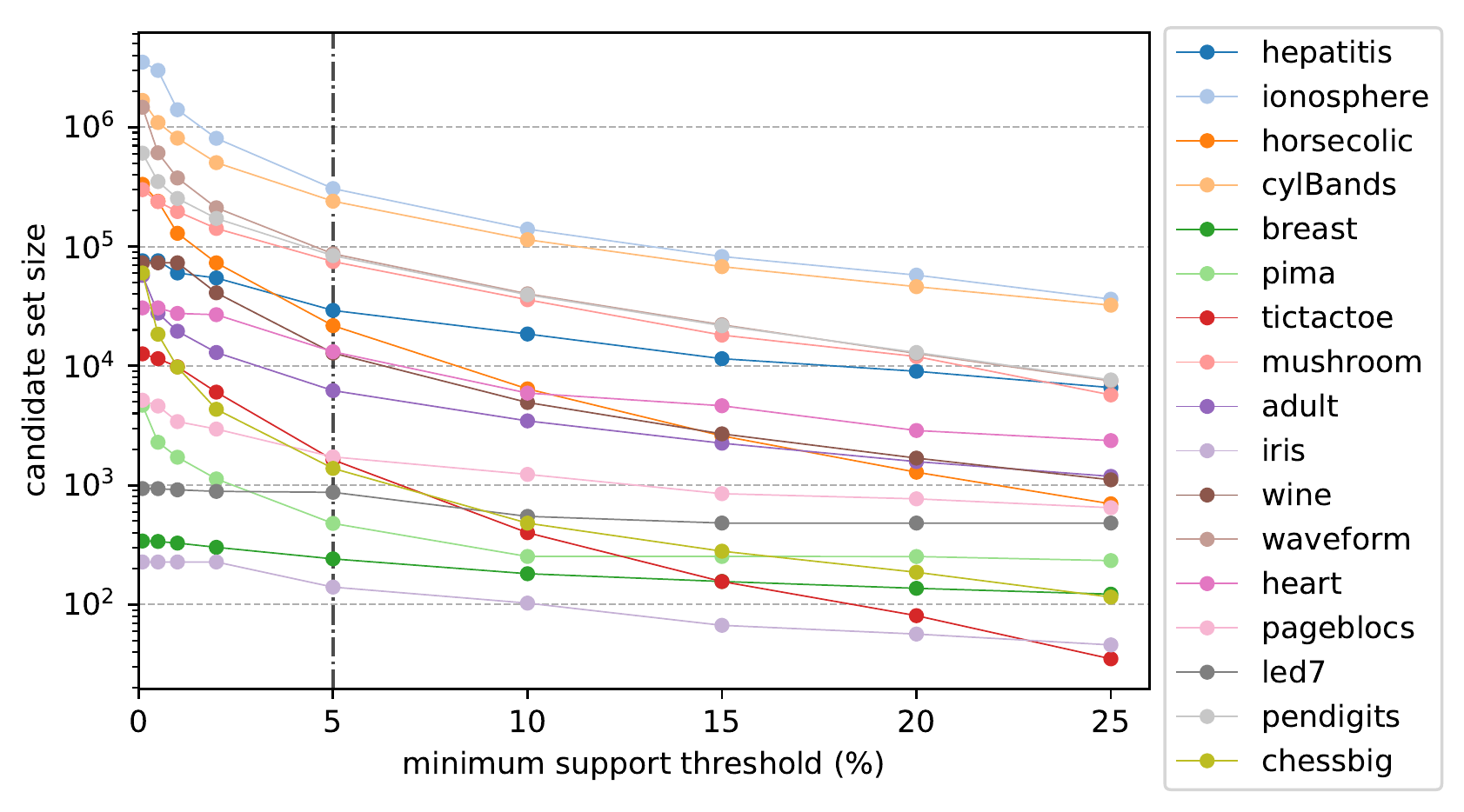}
		\caption{candidate set size}\label{fig:candidatesetsize}
	\end{subfigure}
\end{figure*}
\begin{figure*}[!h]\ContinuedFloat
	\begin{subfigure}{\textwidth}
		\centering
		\includegraphics[width=\linewidth]{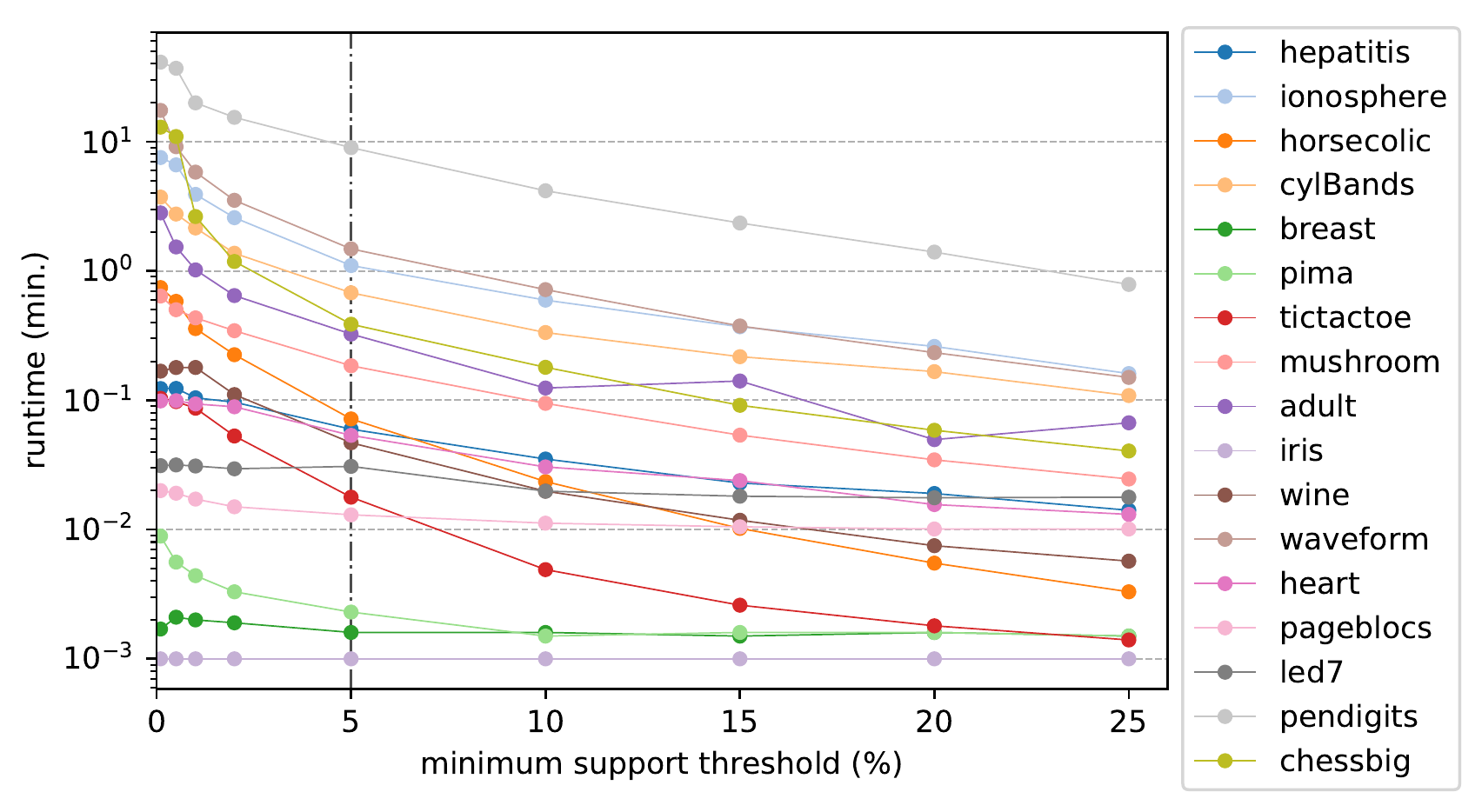}
		\caption{runtime}		\label{fig:candidatetime}
	\end{subfigure}
\end{figure*}

\begin{figure*}[!h]\ContinuedFloat
	\begin{subfigure}{\textwidth}
		\centering
		\includegraphics[width=\linewidth]{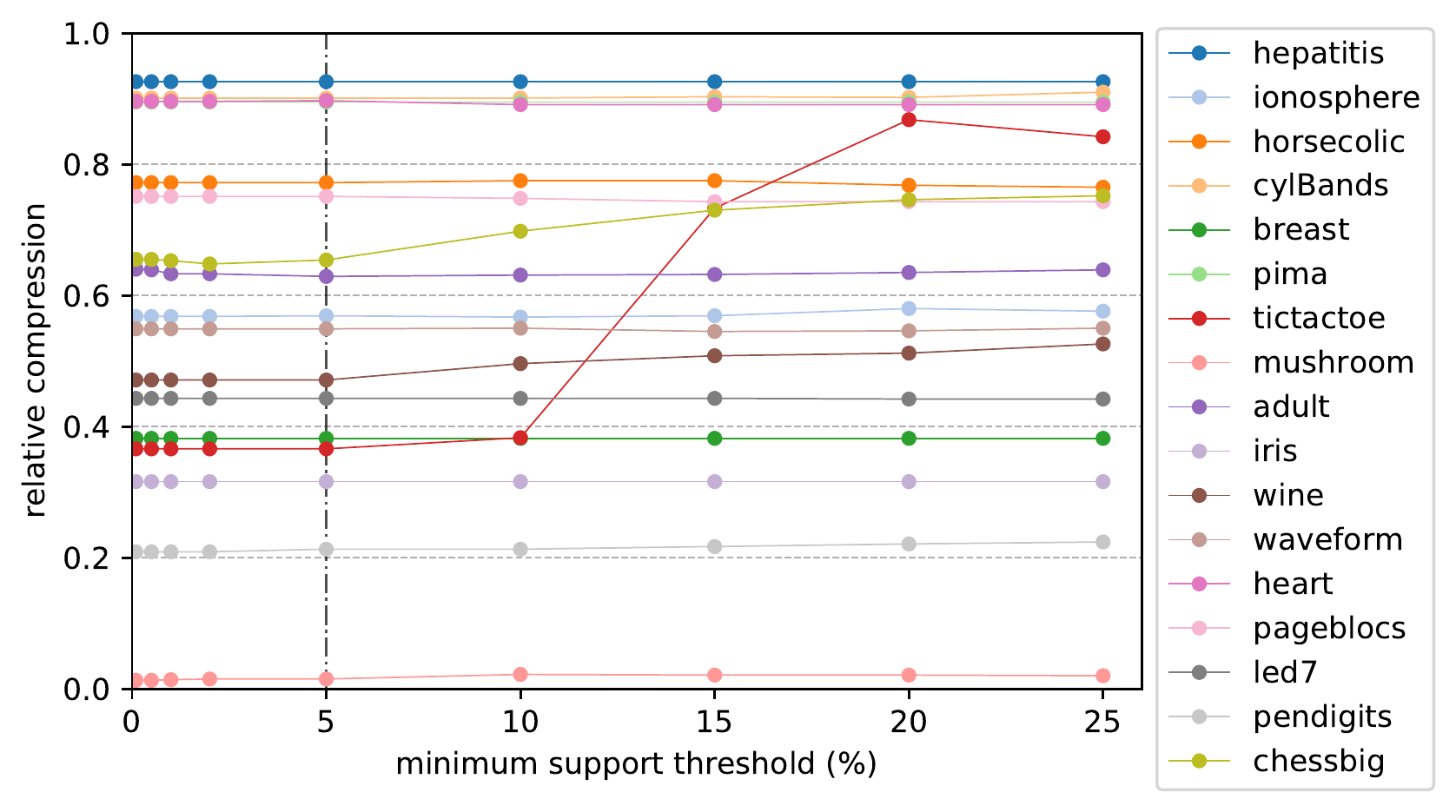}
		\caption{compression in training set}		\label{fig:candidatecompression}
	\end{subfigure}
\end{figure*}
\begin{figure*}[!h]\ContinuedFloat
	\begin{subfigure}{\textwidth}
		\centering
		\includegraphics[width=\linewidth]{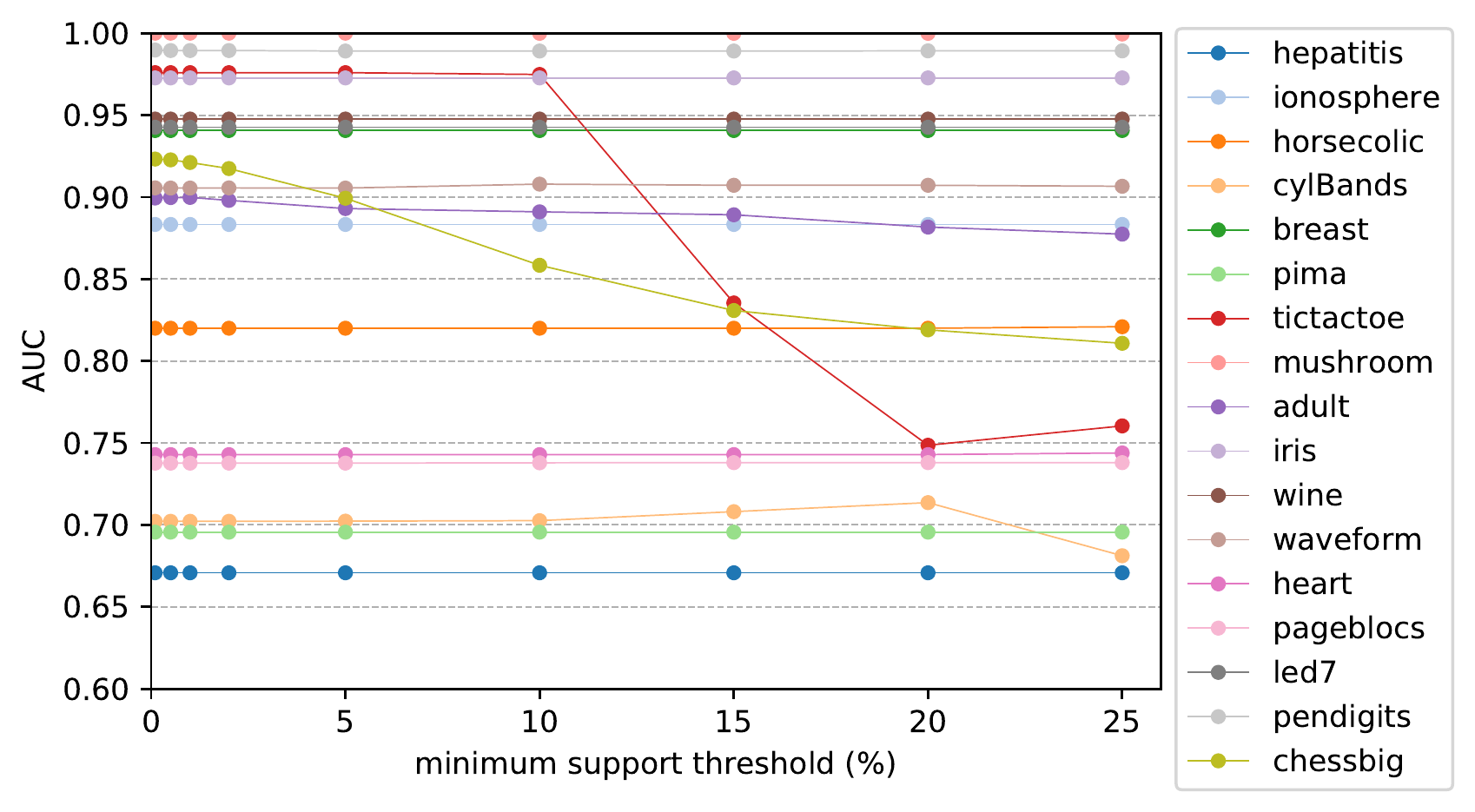}
		\caption{AUC in test set}\label{fig:candidateauc}
	\end{subfigure}
\end{figure*}
\begin{figure*}[!h]\ContinuedFloat
	\begin{subfigure}{\textwidth}
		\centering
		\includegraphics[width=\linewidth]{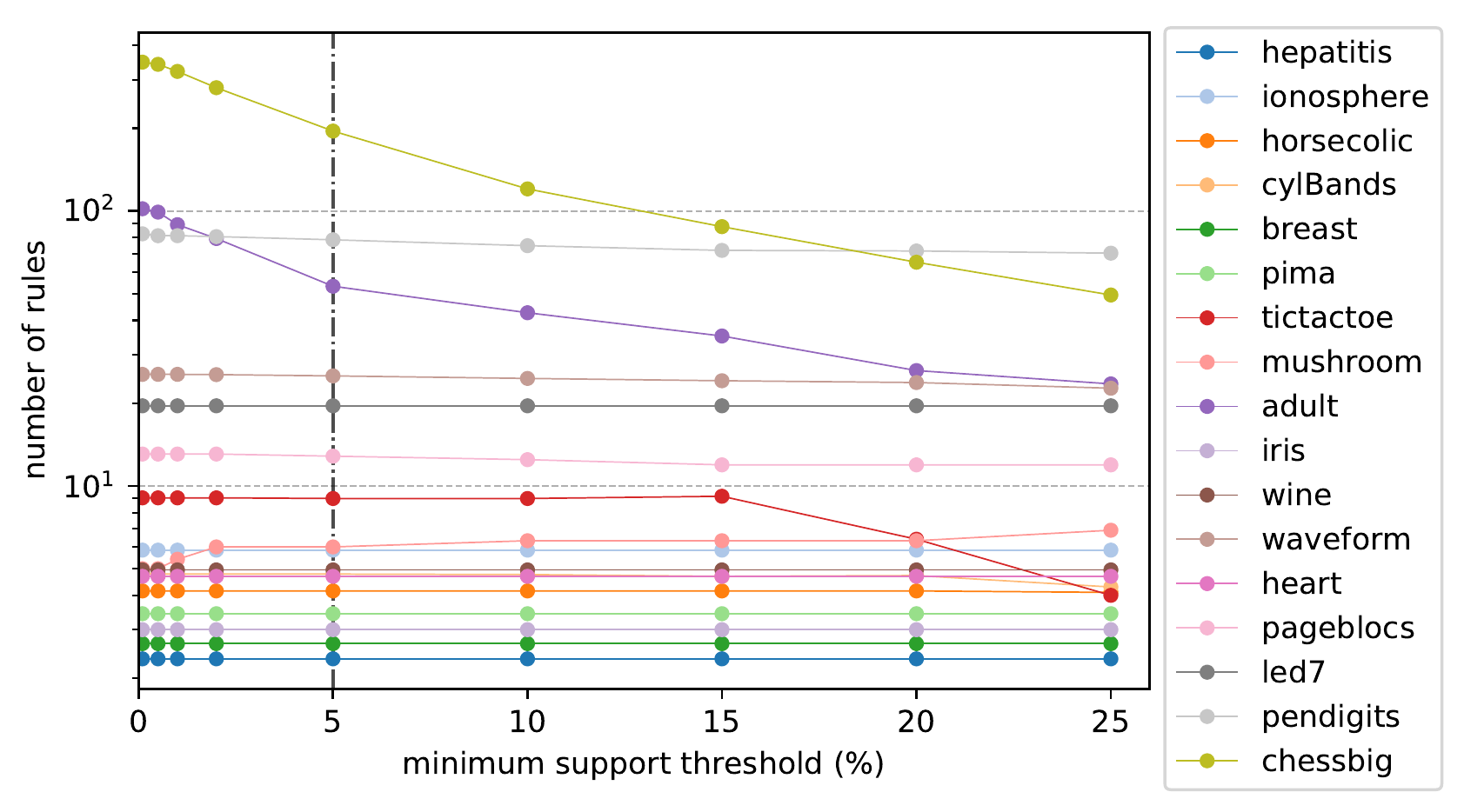}
		\caption{number of rules}		\label{fig:candidaterules}
	\end{subfigure}
	\caption{Influence of the minimum support threshold on \{candidate set size; runtime (in minutes); relative compression on the training set; AUC in the test set; number of rules\}
	for a maximum rule length of $4$ and a minimum support threshold per class of $m_{s} = \{0.1\%,0.5\%,1\%,2\%,5\%,10\%,15\%,20\%,25\% \}$. The values were averaged over $10$ times repeated $10$-fold crossvalidation and each dataset is connected by a line to aid visualization. The vertical dashed line represents the selected minimum support of $5\%$ used to compare against the other algorithms.}
    \label{fig:candidatesetinfluence}
\end{figure*}

The results can be seen in the set of Figures~\ref{fig:candidatesetinfluence}, which show the influence of the candidate set on \Classy{} through: the size of candidates mined in Figure~\ref{fig:candidatesetsize};  runtime in Figure~\ref{fig:candidatetime}; compression on the training set in Figure~\ref{fig:candidatecompression}; AUC in the test set in Figure~\ref{fig:candidateauc}; and the number of rules in a rule list in  Figure~\ref{fig:candidaterules}.

Figure~\ref{fig:candidatesetsize} shows the growth of  of the candidate set size with the minimum support threshold used, and that, as expected, its growth is exponential with the change in minimum support. Figure \ref{fig:candidatetime} shows that in general the runtime increases at a rate similar to the increase in candidate size of figure~\ref{fig:candidatesetsize}. This is in accordance with our analysis of time complexity in Section~\ref{section:complexity}, which tells us that the time complexity of \Classy{} grows proportionally to the dataset size times the candidate set size, thus, given a fixed dataset size it becomes proportional only to the candidate set size.

Figures~\ref{fig:candidatecompression} and \ref{fig:candidateauc} show how \Classy{} performs in classification in terms of compression in the training set and AUC in the test set, respectively. The values for both plots remain constant for most cases, and when a value deteriorates in terms of compression (increase in compression ratio) for smaller candidate sets, it also deteriorates accordingly in terms of AUC (decrease in AUC) in the test set. 
We make two important observations: 1) the minimum in compression is achieved at the minimum support used for $13$ out of $17$ datasets, and in the cases where it does not happen the difference in relative compression is below $1\%$, which tells us that \Classy{} can find a good description of the data using large candidate sets, without too greedily using rules that only cover few instances; 2) the minimum in compression and maximum in AUC are achieved for the same support value for $12$ out $17$ cases, and in the other cases, the difference is usually smaller than $2\%$ in both measures, revealing the robustness of our MDL formulation at obtaining models that generalize well. The main exception is \emph{ionosphere}, where the best AUC is found at the minimum support threshold of $10\%$, while the lowest threshold finds a PRL with $3\%$ lower AUC, without almost any change in compression. This can be explained by the relatively small number of examples of \emph{ionosphere} ($351$ instances) combined with its peculiar structure.

Figure~\ref{fig:candidaterules} shows the number of rules selected based on the candidate set. As expected the number of rules selected only decreases or remains constant with the candidate set, except for \emph{mushroom}. Upon closer inspection, we observe that this is due to the disappearance of a rule with good performance but low coverage from the candidate set, which has to be replaced by a combination of other rules. The cases where many more rules are selected for lower minimum support thresholds, such as \emph{chessbig} and \emph{adult}, have lower compression and higher AUC values for these large number of rules, which makes these selections sustainable.

\subsection{Classification performance}\label{section:classification}
We now compare the classification performance of \Classy{} to Scalable Bayesian Rule Lists (SBRL) \cite{yang2017scalable}, JRip\footnote{\label{note1} \url{https://cran.r-project.org/package=RWeka}}, FURIA\textsuperscript{\ref{note1}}, CART\footnote{\url{https://cran.r-project.org/package=rpart}}, C5.0\footnote{\url{https://cran.r-project.org/package=C50}}, and Support Vector Machines\footnote{\url{https://cran.r-project.org/package=e1071}} (SVM). These methods are state-of-the-art classifiers, and SBRL, CART, C5.0, JRip, and FURIA---a fuzzy unordered rule induction algorithm---are clearly related to our approach. C5.0 is a newer version of C4.5, and JRip is a Java-implementation of RIPPER. 

\Classy{} has no parameters apart from the candidate set, which was generated using FP-growth with $m_s=5\%$ and $l_{max}=4$ for each dataset (as described at the beginning of Section~\ref{sec:experiments}). We tuned CART by selecting the best performing model on the training set from the models generated with the following complexity parameters: $ \{0.001;0.003;0.01; 0.03;$  $0.1\}$. The same was done for C5.0, with confidence factors: $ \{0.05;0.15;0.25;0.35;0.45\} $. The SVM, with radial kernel, was tuned using $3$-fold cross-validation and a grid search on $\gamma =  \{2^{-6:0} \} $ and $ c = \{2^{-4:4} \} $ within the training set. JRip and FURIA were tuned by setting their hyperparameters to $3$ folds, a minimum weight of $2$, and $2$ optimization runs.

SBRL was trained using the guidelines provided by the authors \cite{yang2017scalable}:  the number of chains was set to $25$; iterations to $5000$; $\eta$, representing the average size of patterns in a rule, to $1$; and $\lambda$, representing the average number of rules, to $5$. The algorithm was first run on the training set and then re-run with $\lambda$ changed to the number of rules obtained. In an attempt to follow their guidelines to use around $300$ candidate rules, minimum and maximum itemset length were set to $1$ and $2$ (or $3$ if possible) respectively, while the minimum threshold for the negative and positive classes was set to one of $\{5\%,10\%,15\% \}$. Note that we initially attempted a fair comparison by using the same candidates for SBRL as for \Classy{}, but due to the limitations on the number of rules that SBRL could practically handle this unfortunately turned out to be infeasible.

The results are presented in Table \ref{table:accuracy}. The SVM models achieves the best ranking overall, but they do not belong to the class of interpretable models. 

\Classy{} performs on par with most tree- and rule-based models in terms of accuracy and balanced accuracy, worst than FURIA for these two measures, and better than these in terms of AUCs for multiclass datasets. The better performance of FURIA can be explained by the fact that it uses fuzzy rule sets rather than probabilistic rule lists; this allows for multiple rules to be activated and aggregated for a single classification, which improves predictive performance but makes interpretability less straightforward. This also means that the number of rules and conditions cannot be directly compared: a FURIA rule set consisting of $5$ rules actually translates to up to $32$ unique rules in the rule list setting. Further, FURIA does not provide probabilistic predictions, unlike our approach.

Comparing to other rule list models, such as SBRL and JRip, \Classy{} performs better for most of the measures used. When viewed against the tree-based models, we can see that our method performs on par with CART for most measures and slightly worse than C5.0, except for AUC in the multiclass scenario. Also, as we will show later, C5.0 tends to obtain equivalent rule lists that are much bigger than the ones produced by \Classy, which makes them perform better in general (but not always).

\setlength{\tabcolsep}{0.6pt} \begin{sidewaystable*}\centering																																														
	\scriptsize \ra{0.9} \begin{tabular}{@{}lrrrrrrrrrrrrrrrrrrrrrrr@{}}\toprule																																														
		&	\multicolumn{7}{r}{Accuracy}													& 	& 	\multicolumn{7}{r}{Balanced accuracy}													& 	& 	\multicolumn{7}{r}{AUC}													\\	\cmidrule(l){2-8} \cmidrule(l){10-16} \cmidrule(l){18-24}
		datasets	&{ \scriptsize	\Classy	}&{ \scriptsize	SBRL	}&{ \scriptsize	JRip	}&{ \scriptsize	CART	}&{ \scriptsize	C5.0	}&{ \scriptsize	FURIA	}&{ \scriptsize	SVM}	& 	&{ \scriptsize	\Classy	}&{ \scriptsize	BRL	}&{ \scriptsize	JRip	}&{ \scriptsize	CART	}&{ \scriptsize	C5.0	}&{ \scriptsize	FURIA	}&{ \scriptsize	SVM}	& 	&{ \scriptsize	\Classy	}&{ \scriptsize	BRL	}&{ \scriptsize	JRip	}&{ \scriptsize	CART	}&{ \scriptsize	C5.0	}&{ \scriptsize	FURIA	}&{ \scriptsize	SVM	}\\  	\midrule
		hepatitis 	&$	0.83	$&$	0.78	$&$	0.79	$&$	0.79	$&$	0.79	$&$	0.79	$&$	0.83	$&\phantom{{\tiny-}}	&$	0.66	$&$	0.59	$&$	0.65	$&$	0.67	$&$	0.65	$&$	0.61	$&$	0.70	$&\phantom{{\tiny-}}	&$	0.67	$&$	0.62	$&$	0.64	$&$	0.72	$&$	0.68	$&$	0.70	$&$	0.85	$\\ 	
		ionosphere 	&$	0.89	$&$	0.88	$&$	0.90	$&$	0.91	$&$	0.90	$&$	0.90	$&$	0.92	$& 	&$	0.87	$&$	0.86	$&$	0.89	$&$	0.89	$&$	0.89	$&$	0.88	$&$	0.91	$& 	&$	0.88	$&$	0.88	$&$	0.89	$&$	0.92	$&$	0.92	$&$	0.91	$&$	0.96	$\\ 	
		horsecolic 	&$	0.80	$&$	0.84	$&$	0.81	$&$	0.83	$&$	0.83	$&$	0.83	$&$	0.84	$& 	&$	0.78	$&$	0.82	$&$	0.80	$&$	0.81	$&$	0.82	$&$	0.81	$&$	0.82	$& 	&$	0.82	$&$	0.83	$&$	0.81	$&$	0.85	$&$	0.85	$&$	0.84	$&$	0.88	$\\ 	
		cylBands 	&$	0.70	$&$	0.68	$&$	0.73	$&$	0.73	$&$	0.74	$&$	0.76	$&$	0.81	$& 	&$	0.65	$&$	0.65	$&$	0.72	$&$	0.71	$&$	0.73	$&$	0.74	$&$	0.80	$& 	&$	0.70	$&$	0.73	$&$	0.74	$&$	0.78	$&$	0.78	$&$	0.79	$&$	0.88	$\\ 	
		breast 	&$	0.93	$&$	0.94	$&$	0.93	$&$	0.93	$&$	0.94	$&$	0.94	$&$	0.94	$& 	&$	0.94	$&$	0.95	$&$	0.94	$&$	0.94	$&$	0.95	$&$	0.95	$&$	0.95	$& 	&$	0.94	$&$	0.95	$&$	0.96	$&$	0.95	$&$	0.95	$&$	0.95	$&$	0.96	$\\ 	
		pima 	&$	0.73	$&$	0.74	$&$	0.73	$&$	0.73	$&$	0.73	$&$	0.74	$&$	0.74	$& 	&$	0.66	$&$	0.69	$&$	0.68	$&$	0.67	$&$	0.67	$&$	0.68	$&$	0.67	$& 	&$	0.70	$&$	0.69	$&$	0.68	$&$	0.71	$&$	0.69	$&$	0.68	$&$	0.75	$\\ 	
		tictactoe 	&$	0.98	$&$	0.81	$&$	0.98	$&$	0.92	$&$	0.94	$&$	0.99	$&$	0.99	$& 	&$	0.98	$&$	0.75	$&$	0.97	$&$	0.91	$&$	0.93	$&$	0.98	$&$	0.99	$& 	&$	0.98	$&$	0.86	$&$	0.97	$&$	0.97	$&$	0.98	$&$	1.00	$&$	1.00	$\\ 	
		mushroom 	&$	1.00	$&$	1.00	$&$	1.00	$&$	1.00	$&$	1.00	$&$	1.00	$&$	1.00	$& 	&$	1.00	$&$	1.00	$&$	1.00	$&$	1.00	$&$	1.00	$&$	1.00	$&$	1.00	$& 	&$	1.00	$&$	1.00	$&$	1.00	$&$	1.00	$&$	1.00	$&$	1.00	$&$	1.00	$\\ 	
		adult 	&$	0.85	$&$	0.85	$&$	0.85	$&$	0.85	$&$	0.85	$&$	0.80	$&$	0.86	$& 	&$	0.75	$&$	0.75	$&$	0.74	$&$	0.75	$&$	0.76	$&$	0.74	$&$	0.76	$& 	&$	0.89	$&$	0.88	$&$	0.74	$&$	0.88	$&$	0.87	$&$	0.76	$&$	0.86	$\\ 	\midrule
		rank 	&$	4.4	$&$	4.6	$&$	4.8	$&$	4.9	$&$	4.1	$&$	3.3	$&$	1.9	$& 	&$	4.9	$&$	4.3	$&$	4.4	$&$	4.7	$&$	3.6	$&$	3.8	$&$	2.3	$& 	&$	4.6	$&$	4.8	$&$	5.4	$&$	3.9	$&$	3.8	$&$	3.8	$&$	1.7	$\\ 	\midrule
		iris 	&$	0.95	$&$		$&$	0.94	$&$	0.93	$&$	0.93	$&$	0.93	$&$	0.94	$& 	&$	0.95	$&$		$&$	0.94	$&$	0.93	$&$	0.93	$&$	0.93	$&$	0.94	$& 	&$	0.97	$&$		$&$	0.97	$&$	0.97	$&$	0.97	$&$	0.95	$&$	0.99	$\\ 	
		wine 	&$	0.89	$&$		$&$	0.88	$&$	0.87	$&$	0.89	$&$	0.93	$&$	0.95	$& 	&$	0.90	$&$		$&$	0.89	$&$	0.87	$&$	0.90	$&$	0.92	$&$	0.95	$& 	&$	0.95	$&$		$&$	0.93	$&$	0.93	$&$	0.95	$&$	0.96	$&$	1.00	$\\ 	
		waveform 	&$	0.75	$&$		$&$	0.77	$&$	0.76	$&$	0.76	$&$	0.78	$&$	0.80	$& 	&$	0.75	$&$		$&$	0.77	$&$	0.76	$&$	0.76	$&$	0.78	$&$	0.80	$& 	&$	0.91	$&$		$&$	0.87	$&$	0.90	$&$	0.90	$&$	0.86	$&$	0.94	$\\ 	
		heart 	&$	0.56	$&$		$&$	0.54	$&$	0.57	$&$	0.54	$&$	0.57	$&$	0.59	$& 	&$	0.30	$&$		$&$	0.22	$&$	0.31	$&$	0.30	$&$	0.27	$&$	0.30	$& 	&$	0.74	$&$		$&$	0.54	$&$	0.76	$&$	0.72	$&$	0.69	$&$	0.84	$\\ 	
		pageblocs 	&$	0.93	$&$		$&$	0.93	$&$	0.93	$&$	0.92	$&$	0.92	$&$	0.93	$& 	&$	0.51	$&$		$&$	0.52	$&$	0.52	$&$	0.48	$&$	0.52	$&$	0.52	$& 	&$	0.74	$&$		$&$	0.72	$&$	0.74	$&$	0.69	$&$	0.73	$&$	0.72	$\\ 	
		led7 	&$	0.74	$&$		$&$	0.72	$&$	0.75	$&$	0.75	$&$	0.74	$&$	0.76	$& 	&$	0.74	$&$		$&$	0.72	$&$	0.75	$&$	0.75	$&$	0.74	$&$	0.76	$& 	&$	0.94	$&$		$&$	0.92	$&$	0.94	$&$	0.94	$&$	0.88	$&$	0.95	$\\ 	
		pendigits 	&$	0.92	$&$		$&$	0.95	$&$	0.92	$&$	0.96	$&$	0.97	$&$	0.98	$& 	&$	0.92	$&$		$&$	0.95	$&$	0.92	$&$	0.96	$&$	0.97	$&$	0.99	$& 	&$	0.99	$&$		$&$	0.98	$&$	0.99	$&$	1.00	$&$	0.99	$&$	1.00	$\\ 	
		chessbig 	&$	0.50	$&$		$&$	0.52	$&$	0.49	$&$	0.78	$&$	0.62	$&$	0.93	$& 	&$	0.44	$&$		$&$	0.61	$&$	0.41	$&$	0.81	$&$	0.67	$&$	0.92	$& 	&$	0.90	$&$		$&$	0.81	$&$	0.87	$&$	0.96	$&$	0.67	$&$	1.00	$\\ 	\midrule
		rank 	&$	4.0	$&$		$&$	3.4	$&$	3.2	$&$	2.6	$&$	1.6	$&$	1.1	$& 	&$	4.1	$&$		$&$	4.4	$&$	4.1	$&$	4.1	$&$	3.2	$&$	1.3	$& 	&$	2.6	$&$		$&$	5.1	$&$	3.8	$&$	3.4	$&$	4.6	$&$	1.5	$\\ 	
		\bottomrule						
	\end{tabular}																								
	\caption{Classification performance average results (10 times repeated 10-fold crossvalidation) measured through \{accuracy; balanced accuracy; Area Under the ROC Curve (AUC) (weighted AUC for multiclass datasets)\}, per dataset  \{binary;multiclass\} for each algorithm. At the bottom of the binary and multiclass datasets the average rank of the algorithm for each is presented. The rank of $1$ is given for the best value (highest classification performance) and $7$ and $6$ for the worst in binary and multiclass, respectively. Note that SBRL cannot do multiclass classification, hence only being present in the binary ranking and the blank spaces.}\label{table:accuracy}																																														
\end{sidewaystable*}																																							
									
\subsection{Interpretability}
The results are shown in Table~\ref{table:interpretability}, where we use AUC, the number of rules, and the number of conditions to compare the trade-off between AUC and model complexity of the tree- and rule-based models. Note that we choose AUC for predictive performance as it agrees with our goal of using the probabilities output of \Classy{} to explain the decisions made. Also note that we intentionally removed FURIA from the rankings of the number of rules and conditions as its models are rule sets---not rule lists. %This means that if a model has for example $6$ independent rules (as in the case of horsecolic) they all can interact with each other and make in the maximum an equivalent rule list of $63$ rules. 

For binary datasets \Classy{} is in a middle ranking, better than SBRL and JRip, and worst than CART, C5.0 nd FURIA. On the other hand, in multiclass datasets it achieves a much lower (=better) ranking than all the other algorithms. 

\Classy{} tends to find more compact models, with similar number of rules and fewer logical conditions in total, than C5.0, CART, and JRip, that are as accurate or better than these. This can be clearly seen by its average rank of $2$ and $1.9$ for rules and $1.7$ and $1.5$ for the total number of conditions, for binary and multiclass datasets respectively. It also can be seen that for most datasets it obtained the lowest number of conditions of all tree- and rule-based classifiers. Although SBRL also finds very compact rule lists, with small number of rules and conditions, the low variance between the reported values for the different datasets suggests that this strongly depends on the hyperparameter settings, which penalize too strongly the number of rules not around the user defined expected average number of rules. Indeed, the compact rule lists exhibit subpar classification performance for some datasets (i.e., \emph{hepatitis} and \emph{tictactoe}). This suggests that without additional (computation-intensive) tuning of these hyperparameters, the recommended procedure for SBRL may lead to underfitting. As expected, C5.0, with its tendency to maximize the classification performance as much as possible, tends to create overgrown models, such as the almost $3000$ rules for \emph{chessbig}, that do not necessarily generalize well, such is the case in \emph{adult}, where it obtained the same number of rules as \Classy{} but with a $2\%$ lower AUC, and for \emph{pendigits} were it obtained a number of rules around $4$ times higher than \Classy{} and CART for the same performance.

\setlength{\tabcolsep}{1.5pt}
\begin{sidewaystable*}\centering															
	\scriptsize \ra{0.87} \begin{tabular}{@{}lrrrrrrrrrrrrrrrrrrrr@{}}\toprule																																												
		&	\multicolumn{6}{r}{AUC}												& 	& 	\multicolumn{6}{r}{Number of rules}												& 	& 	\multicolumn{6}{r}{Number of conditions}													\\	\cmidrule(l){2-7} \cmidrule(l){9-14} \cmidrule(l){16-21} 
		datasets	&{ \scriptsize	\Classy	}&{ \scriptsize	SBRL	}&{ \scriptsize	JRip	}&{ \scriptsize	CART	}&{ \scriptsize	C5.0	}&{ \scriptsize	FURIA}		& 	&{ \scriptsize	\Classy	}&{ \scriptsize	SBRL	}&{ \scriptsize	JRip	}&{ \scriptsize	CART	}&{ \scriptsize	C5.0	}&{ \scriptsize	FURIA}		& 	&{ \scriptsize	\Classy	}&{ \scriptsize	SBRL	}&{ \scriptsize	JRip	}&{ \scriptsize	CART	}&{ \scriptsize	C5.0	}&{ \scriptsize	FURIA			}\\  	\hline
		hepatitis 	&$	0.67	$&$	0.62	$&$	0.64	$&$	0.72	$&$	0.68	$&$	0.70		$&\phantom{{\tiny-}}	&$	2	$&$	2	$&$	3	$&$	4	$&$	9	$&$	4	^{*}	$& \phantom{{\tiny-}}	&$	1	$&$	2	$&$	5	$&$	6	$&$	38	$&$	7	^{*}		$\\ 	
		ionosphere 	&$	0.88	$&$	0.88	$&$	0.89	$&$	0.92	$&$	0.92	$&$	0.91		$& 	&$	6	$&$	3	$&$	6	$&$	5	$&$	12	$&$	7	^{*}	$& 	&$	4	$&$	4	$&$	9	$&$	10	$&$	58	$&$	12	^{*}		$\\ 	
		horsecolic 	&$	0.82	$&$	0.83	$&$	0.81	$&$	0.85	$&$	0.85	$&$	0.84		$& 	&$	4	$&$	2	$&$	4	$&$	6	$&$	16	$&$	6	^{*}	$& 	&$	3	$&$	2	$&$	7	$&$	10	$&$	76	$&$	9	^{*}		$\\ 	
		cylBands 	&$	0.70	$&$	0.73	$&$	0.74	$&$	0.78	$&$	0.78	$&$	0.79		$& 	&$	5	$&$	3	$&$	7	$&$	20	$&$	59	$&$	11	^{*}	$& 	&$	4	$&$	4	$&$	17	$&$	111	$&$	897	$&$	18	^{*}		$\\ 	
		breast 	&$	0.94	$&$	0.95	$&$	0.96	$&$	0.95	$&$	0.95	$&$	0.95		$& 	&$	3	$&$	3	$&$	4	$&$	5	$&$	6	$&$	5	^{*}	$& 	&$	3	$&$	5	$&$	11	$&$	13	$&$	15	$&$	13	^{*}		$\\ 	
		pima 	&$	0.70	$&$	0.69	$&$	0.68	$&$	0.71	$&$	0.69	$&$	0.68		$& 	&$	3	$&$	2	$&$	3	$&$	10	$&$	11	$&$	2	^{*}	$& 	&$	3	$&$	3	$&$	3	$&$	23	$&$	47	$&$	2	^{*}		$\\ 	
		tictactoe 	&$	0.98	$&$	0.86	$&$	0.97	$&$	0.97	$&$	0.98	$&$	1.00		$& 	&$	9	$&$	6	$&$	10	$&$	24	$&$	42	$&$	10	^{*}	$& 	&$	21	$&$	15	$&$	27	$&$	66	$&$	254	$&$	17	^{*}		$\\ 	
		mushroom 	&$	1.00	$&$	1.00	$&$	1.00	$&$	1.00	$&$	1.00	$&$	1.00		$& 	&$	6	$&$	5	$&$	5	$&$	8	$&$	9	$&$	8	^{*}	$& 	&$	7	$&$	8	$&$	7	$&$	25	$&$	35	$&$	17	^{*}		$\\ 	
		adult 	&$	0.89	$&$	0.88	$&$	0.74	$&$	0.88	$&$	0.87	$&$	0.76		$& 	&$	53	$&$	13	$&$	17	$&$	22	$&$	52	$&$	21	^{*}	$& 	&$	114	$&$	25	$&$	81	$&$	106	$&$	630	$&$	34	^{*}		$\\ 	\midrule
		rank 	&$	3.8	$&$	4.0	$&$	4.4	$&$	3.0	$&$	2.9	$&$	2.9		$& 	&$	2.6	$&$	1.1	$&$	2.8	$&$	3.7	$&$	4.9	$&$	\phantom{2.3}	^{*}	$& 	&$	1.7	$&$	1.7	$&$	2.7	$&$	3.9	$&$	5.0	$&$	\phantom{2.3}	^{*}		$\\ 	\midrule
		iris 	&$	0.97	$&$		$&$	0.97	$&$	0.97	$&$	0.97	$&$	0.95		$& 	&$	3	$&$		$&$	3	$&$	3	$&$	3	$&$	2	^{*}	$& 	&$	2	$&$		$&$	2	$&$	3	$&$	3	$&$	3	^{*}		$\\ 	
		wine 	&$	0.95	$&$		$&$	0.93	$&$	0.93	$&$	0.95	$&$	0.96		$& 	&$	5	$&$		$&$	5	$&$	5	$&$	8	$&$	4	^{*}	$& 	&$	4	$&$		$&$	7	$&$	6	$&$	24	$&$	6	^{*}		$\\ 	
		waveform 	&$	0.91	$&$		$&$	0.87	$&$	0.90	$&$	0.90	$&$	0.86		$& 	&$	25	$&$		$&$	23	$&$	46	$&$	81	$&$	15	^{*}	$& 	&$	65	$&$		$&$	108	$&$	125	$&$	683	$&$	33	^{*}		$\\ 	
		heart 	&$	0.74	$&$		$&$	0.54	$&$	0.76	$&$	0.72	$&$	0.69		$& 	&$	5	$&$		$&$	3	$&$	11	$&$	39	$&$	2	^{*}	$& 	&$	4	$&$		$&$	6	$&$	28	$&$	299	$&$	2	^{*}		$\\ 	
		pageblocs 	&$	0.74	$&$		$&$	0.72	$&$	0.74	$&$	0.69	$&$	0.73		$& 	&$	13	$&$		$&$	8	$&$	10	$&$	9	$&$	3	^{*}	$& 	&$	13	$&$		$&$	9	$&$	20	$&$	37	$&$	4	^{*}		$\\ 	
		led7 	&$	0.94	$&$		$&$	0.92	$&$	0.94	$&$	0.94	$&$	0.88		$& 	&$	20	$&$		$&$	19	$&$	29	$&$	28	$&$	3	^{*}	$& 	&$	46	$&$		$&$	74	$&$	43	$&$	138	$&$	6	^{*}		$\\ 	
		pendigits 	&$	0.99	$&$		$&$	0.98	$&$	0.99	$&$	1.00	$&$	0.99		$& 	&$	78	$&$		$&$	107	$&$	66	$&$	266	$&$	74	^{*}	$& 	&$	221	$&$		$&$	439	$&$	30	$&$	2934	$&$	133	^{*}		$\\ 	
		chessbig 	&$	0.90	$&$		$&$	0.81	$&$	0.87	$&$	0.96	$&$	0.67		$& 	&$	195	$&$		$&$	418	$&$	118	$&$	2874	$&$	281	^{*}	$& 	&$	483	$&$		$&$	2688	$&$	25	$&$	48826	$&$	803	^{*}		$\\ 	\midrule
		rank 	&$	1.8	$&$		$&$	4.3	$&$	2.9	$&$	2.4	$&$	3.8		$& 	&$	2.3	$&$		$&$	2.0	$&$	2.2	$&$	3.5	$&$	\phantom{2.3}	^{*}	$& 	&$	1.5	$&$		$&$	2.4	$&$	2.1	$&$	4.0	$&$	\phantom{2.3}	^{*}		$\\ 	
		\bottomrule					
	\end{tabular}
	\caption{Interpretability performance average results (10 times repeated 10-fold crossvalidation) of tree- and rule-based models measured through \{Area Under the ROC Curve (AUC) (weighted AUC for multiclass datasets); number of rules; number of conditions \}, per dataset  \{binary; multiclass\} for each algorithm. Rank gives the average rank of each algorithm for binary and multiclass datasets. The rank of $1$ is given for the best value (highest AUC or lowest  number of rules/conditions). Note that SBRL cannot do multiclass classification, hence only being present in the binary ranking and the blank spaces. \newline $*$ The number of rules and conditions used by FURIA are presented as a reference, as they are not directly comparable to those of the other methods as they form rule sets (and cannot be trivially translated to rule lists). FURIA was therefore also not included in the rankings of those criteria. }\label{table:interpretability}								
\end{sidewaystable*}

\subsection{Statistical significance testing}
To analyze whether the results Tables~\ref{table:accuracy} and \ref{table:interpretability} are statistically different \cite{demvsar2006statistical}, we use two non-parametric multiple hypothesis tests, namely Friedman's test \cite{friedman1937use} and Iman and Davenport's test\cite{iman1980approximations}, on the rankings of the algorithms. 

The results can be seen in the left side of Table~\ref{table:Friedman}, which divides the datasets into two groups, for binary and multiclass datasets respectively. The results show that there are significant differences for most measures (significance level $0.05$). The only exceptions are balanced accuracy in the binary case, AUC of rule-based models in the binary case, and the number of rules for the multi-class case.\\

For those cases where the null hypothesis---stating that the algorithms perform on par---is rejected we proceed with a post-hoc Holm's test \cite{holm1979simple} for pairwise comparisons with \Classy{} as control algorithm. 

The results of these pairwise comparisons can be seen in the right side of Table~\ref{table:Friedman}. For most of these tests the null hypothesis---stating that \Classy{} and its competitor perform on par---cannot be rejected. This can be mostly explained by the relatively small number of datasets; the power of the tests is not very high. We therefore cannot draw strong conclusions from these results, but this is not necessarily a negative outcome: we aimed at showing that \Classy{} performs as well as other rule- and tree-based algorithms while obtaining simpler models. The results show that \Classy{} does use significantly fewer conditions than C5.0 for both multiclass and binary datasets, and than CART for the binary case. Also, as expected the SVM obtained always better results than \Classy{} except for multiclass AUC. FURIA was better in terms of accuracy but worse in terms of AUC.

\begin{sidewaystable*}\centering																	
	\scriptsize \ra{1.3} \begin{tabular}{@{}llrrrrrrrrrrrrrrrrrrrr@{}}\toprule																																												
		&		&		&	\multicolumn{6}{r}{Difference tests}											&	&	\multicolumn{12}{r}{Holm's post-hoc procedure (\Classy{} as control) }																								\\	\cmidrule(l){3-9} \cmidrule(l){11-22}
		&		&		&	\multicolumn{3}{r}{Friedman}					&	\multicolumn{3}{r}{\phantom{a} Iman and Davenport}					&\phantom{ab}	&	\multicolumn{2}{r}{SBRL}			&	\multicolumn{2}{r}{JRip}			&	\multicolumn{2}{r}{CART}			&	\multicolumn{2}{r}{C5.0}			&	\multicolumn{2}{r}{FURIA}			&	\multicolumn{2}{r}{SVM}				\\	\cmidrule(l){4-6} \cmidrule(l){7-9} \cmidrule(l){11-12} \cmidrule(l){13-14} \cmidrule(l){15-16} \cmidrule(l){17-18} \cmidrule(l){19-20} \cmidrule(l){21-22}
		Measures	&	Classes	&	$k$	&	$\chi^2$	&$	p	$&\phantom{a}	$\mathcal{H}_0$	&	$F$	&$	p	$&\phantom{a}	$\mathcal{H}_0$	&	&$	p	$&\phantom{a}	$\mathcal{H}_0$	&$	p	$&\phantom{a}	$\mathcal{H}_0$	&$	p	$&\phantom{a}	$\mathcal{H}_0$	&$	p	$&\phantom{a}	$\mathcal{H}_0$	&$	p	$&\phantom{a}	$\mathcal{H}_0$	&$	p	$&	\phantom{a} $\mathcal{H}_0$		\\	\midrule
		Acc	&	binary	&$	7	$&$\phantom{a}	13.36	$&$	0.038	$&	\phantom{a}\textbf{R}	&$	2.63	$&$	0.028	$&	\phantom{a}\textbf{R}	&	&\phantom{a}$	0.827	$&	---	&$	0.703	$&	---	&$	0.585	$&	---	&$	0.743	$&	---	&$	0.300	$&	---	&$	0.014	$&	\phantom{a}\textbf{R}		\\	
		&	multi	&$	6	$&$	17.34	$&$	0.004	$&	\phantom{a}\textbf{R}	&$	5.36	$&$	<\! 0.001	$&	\phantom{a}\textbf{R}	&	&$		$&		&\phantom{a}$	0.504	$&	---	&\phantom{a}$	0.385	$&	---	&$	0.142	$&	---	&$	0.009	$&	\phantom{a}\textbf{R}	&\phantom{a}$	0.002	$&	\phantom{a}\textbf{R}		\\	
		bAcc	&	binary	&$	7	$&$	8.94	$&$	0.177	$&	---	&$	1.59	$&$	0.171	$&	---	&	&		&		&		&		&		&		&		&		&		&		&		&			\\	
		&	multi	&$	6	$&$	15.71	$&$	0.008	$&	\phantom{a}\textbf{R}	&$	4.53	$&$	0.003	$&	\phantom{a}\textbf{R}	&	&$		$&		&$	0.738	$&	---	&$	1.000	$&	---	&$	1.000	$&	---	&$	0.350	$&	---	&$	0.003	$&	\phantom{a}\textbf{R}		\\	
		$AUC_{all}$	&	binary	&$	7	$&$	16.00	$&$	0.014	$&	\phantom{a}\textbf{R}	&$	3.37	$&$	0.007	$&	\phantom{a}\textbf{R}	&	&$	0.827	$&	---	&$	0.445	$&	---	&$	0.300	$&	---	&$	0.413	$&	---	&$	0.413	$&	---	&$	0.005	$&	\phantom{a}\textbf{R}		\\	
		&	multi	&$	6	$&$	20.00	$&$	0.001	$&	\phantom{a}\textbf{R}	&$	7.00	$&$\phantom{a}	<\! 0.001	$&	\phantom{a}\textbf{R}	&	&$		$&		&$	0.008	$&	\phantom{a}\textbf{R}	&$	0.229	$&	---	&$	0.423	$&	---	&$\phantom{a}	0.033	$&	---	&$	0.229	$&	---		\\	
		$AUC_{rules}$	&	binary	&$	6	$&$	5.70	$&$	0.337	$&	---	&$	1.16	$&$	0.346	$&	---	&	&		&		&		&		&		&		&		&		&		&		&		&			\\	
		&	multi	&$	5	$&$	13.10	$&$	0.011	$&	\phantom{a}\textbf{R}	&$	4.85	$&$	0.004	$&	\phantom{a}\textbf{R}	&	&$		$&		&$	0.002	$&	\phantom{a}\textbf{R}	&$	0.155	$&	---	&$	0.429	$&	---	&$	0.011	$&	\phantom{a}\textbf{R}	&$		$&			\\	
		Rules	&	binary	&$	5	$&$	28.18	$&$\phantom{a}	<\! 0.001	$&	\phantom{a}\textbf{R}	&$	28.82	$&$	<\! 0.001	$&	\phantom{a}\textbf{R}	&	&$	0.053	$&	---	&$	0.766	$&	---	&$	0.136	$&	---	&$	0.002	$&	\phantom{a}\textbf{R}	&$		$&		&$		$&			\\	
		&	multi	&$	4	$&$	6.64	$&$	0.084	$&	---	&$	2.68	$&$	0.073	$&	---	&	&		&		&		&		&		&		&		&		&		&		&		&			\\	
		Conditions	&	binary	&$	5	$&$	29.40	$&$	<\! 0.001	$&	\phantom{a}\textbf{R}	&$	35.64	$&$	<\! 0.001	$&	\phantom{a}\textbf{R}	&	&$	1.000	$&	---	&$	0.205	$&	---	&$	0.011	$&	\phantom{a}\textbf{R}	&\phantom{a}$	<\! 0.001	$&	\phantom{a}\textbf{R}	&$		$&		&$		$&			\\	
		&	multi	&$	4	$&$	16.35	$&$	0.001	$&	\phantom{a}\textbf{R}	&$	14.96	$&$	<\! 0.001	$&	\phantom{a}\textbf{R}	&	&$		$&		&$	0.175	$&	---	&$	0.333	$&	---	&$	<\! 0.001	$&	\phantom{a}\textbf{R}	&$		$&		&$		$&			\\	
		\bottomrule																																												
	\end{tabular}																																												
	\caption{Statistical significance testing of differences between the algorithms. Results for Friedman ($\chi^2$ statistic), and Iman and Davenport ($F$ statistic)  tests for all the measures presented in Table~\ref{table:accuracy} and \ref{table:interpretability}  for binary and multiclass  datasets with a significance level of $0.05$. In case the null hypothesis for the differences is rejected, the post-hoc Holm's procedure is used for pairwise comparisons with \Classy{} as control. $AUC_{all}$ is the AUC comparison with all algorithms (SVM included) of Table~\ref{table:accuracy}  and   $AUC_{rules}$ is the AUC comparison of all tree- and rule-based algorithms (SVM excluded) of Table~\ref{table:interpretability}.  $p$ represents the p-value obtained for each specific test, \textbf{R} the rejection of $\mathcal{H}_0$---null hypothesis. Note that not all algorithms can be tested for all measures, thus $k$ shows the number of algorithms tested for each measure.  } \label{table:Friedman}																																												
\end{sidewaystable*}																																							
\subsection{Overfitting}

To study overfitting, we compared the averages of the absolute difference between the AUC values in the training and test set over $10$ times repeated $10$-folds for each algorithm. 
The results can be seen in Table~\ref{table:overfitting}. In general \Classy{}, together with SVM, seem to be the most consistent algorithms in obtaining the lowest values. The usual performance of \Classy{} is $5\%$ or lower, $12$ out of $17$ times, except in the case of \emph{hepatitis} were it got $13\%$, which was the best value after SVM. SBRL is very consistent, clearly achieving the lowest values for binary datasets, however this can be explained by its more conservative choice of rules and thus lower AUC on the test set as shown in Table~\ref{table:accuracy}. Comparing with all rule- and tree-based models, \Classy{} obtained the lowest ranking for multiclass datasets, being, from these ones, the algorithm that less overfits overall.

\setlength{\tabcolsep}{1.4pt}  \begin{table}[h!]\centering																	
	\scriptsize \ra{1.0} \begin{tabular}{@{}lrrrrrrrr@{}} \toprule																	
		&	\multicolumn{7}{r}{$|AUC_{train}-AUC_{test}|$}														\\	\cmidrule(l){2-8} 
		datasets	&	\Classy	&	SBRL	&	JRip	&	CART	&	C5.0	&	FURIA	&	SVM		\\	\midrule
		hepatitis 	&$	0.13	$&$	0.16	$&$	0.19	$&$	0.14	$&$	0.21	$&$	0.22	$&$	0.13	$	\\	
		ionosphere 	&$	0.06	$&$	0.05	$&$	0.07	$&$	0.04	$&$	0.05	$&$	0.08	$&$	0.04	$	\\	
		horsecolic 	&$	0.06	$&$	0.06	$&$	0.08	$&$	0.06	$&$	0.09	$&$	0.08	$&$	0.10	$	\\	
		cylBands 	&$	0.07	$&$	0.06	$&$	0.09	$&$	0.11	$&$	0.18	$&$	0.13	$&$	0.12	$	\\	
		breast 	&$	0.02	$&$	0.02	$&$	0.02	$&$	0.02	$&$	0.02	$&$	0.02	$&$	0.02	$	\\	
		pima 	&$	0.06	$&$	0.05	$&$	0.05	$&$	0.06	$&$	0.07	$&$	0.05	$&$	0.05	$	\\	
		tictactoe 	&$	0.01	$&$	0.03	$&$	0.01	$&$	0.02	$&$	0.02	$&$	0.00	$&$	0.00	$	\\	
		mushroom 	&$	0.00	$&$	0.00	$&$	0.00	$&$	0.00	$&$	0.00	$&$	0.00	$&$	0.00	$	\\	
		adult 	&$	0.01	$&$	0.00	$&$	0.01	$&$	0.00	$&$	0.01	$&$	0.01	$&$	0.03	$	\\	\midrule
		rank	&$	3.6	$&$	2.7	$&$	4.4	$&$	4.2	$&$	5.2	$&$	4.3	$&$	3.6	$	\\	\midrule
		iris 	&$	0.02	$&$		$&$	0.02	$&$	0.02	$&$	0.02	$&$	0.04	$&$	0.01	$	\\	
		wine 	&$	0.03	$&$		$&$	0.05	$&$	0.04	$&$	0.04	$&$	0.03	$&$	0.00	$	\\	
		waveform 	&$	0.02	$&$		$&$	0.02	$&$	0.02	$&$	0.03	$&$	0.02	$&$	0.02	$	\\	
		heart 	&$	0.05	$&$		$&$	0.04	$&$	0.07	$&$	0.15	$&$	0.04	$&$	0.06	$	\\	
		pageblocs 	&$	0.00	$&$		$&$	0.00	$&$	0.00	$&$	0.00	$&$	0.00	$&$	0.00	$	\\	
		led7 	&$	0.01	$&$		$&$	0.01	$&$	0.01	$&$	0.01	$&$	0.01	$&$	0.01	$	\\	
		pendigits 	&$	0.00	$&$		$&$	0.01	$&$	0.00	$&$	0.00	$&$	0.01	$&$	0.00	$	\\	
		chessbig 	&$	0.00	$&$		$&$	0.02	$&$	0.00	$&$	0.02	$&$	0.00	$&$	0.00	$	\\	\midrule
		rank	&$	2.4	$&$		$&$	4.8	$&$	4.4	$&$	4.1	$&$	3.6	$&$	1.8	$	\\	
		\bottomrule																	
	\end{tabular}																	
	\caption{Overfitting average results (10 times repeated 10-fold crossvalidation) using the absolute difference between AUC performance in training and test sets as measure, per fold, for each algorithm and each dataset. Rank gives the average rank of each algorithm for binary and multiclass datasets. Note that SBRL does not have a values for multiclass datasets. }\label{table:overfitting}																	
\end{table}

\subsection{Runtime}													
All runtimes are averages over ten times repetitions of ten folds, run on a $64$-bit Windows Server $2012R2$, with Intel Xeon E5-2630v3 CPU at 2.4GHz and 512GB RAM. Runtimes include parameter tuning where applicable, and candidate mining for \Classy{} and SBRL.

\begin{figure}[!h]
	\centering
	\includegraphics[clip,width=1\textwidth]{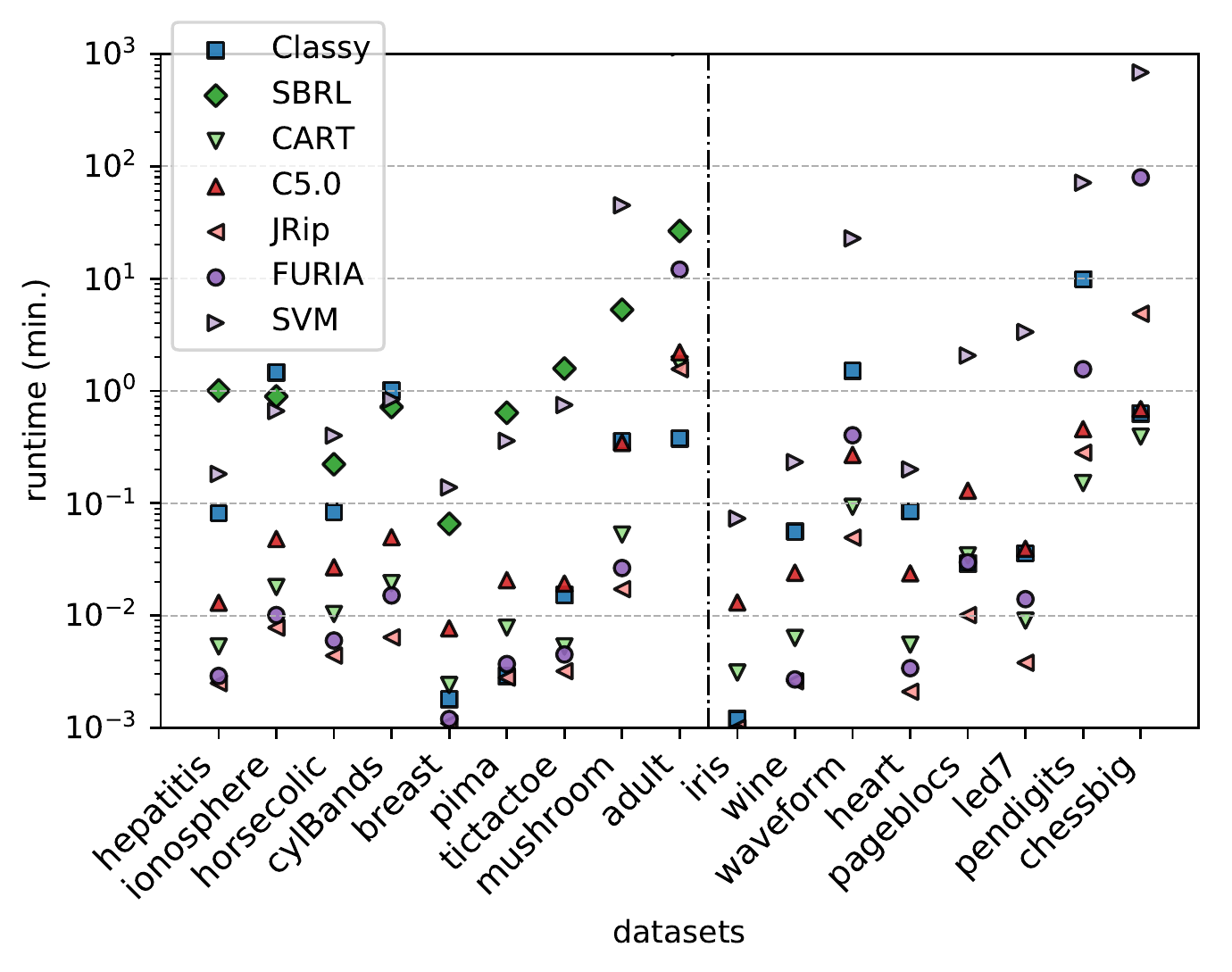}
	\caption{Average runtime per fold in minutes for each algorithm and each dataset. The datasets are ordered first by the number classes and then by number of samples (ascending). The vertical dashed line separates binary (to the left) from multiclass datasets. Note that SBRL does not have a runtime for multiclass datasets.}
	\label{fig:runtime}
\end{figure}

The results are depicted in Figure~\ref{fig:runtime}. CART, C5.0, JRip and FURIA are the fastest, with most runtimes under $1$ minute with \Classy{} being at maximum one order of magnitude slower. Comparing to SBRL, \Classy{} is $10$ times faster, even though it considers around $100$ times more candidates than this and performs better in terms of AUC. The worst runtimes were obtained for SVM, due to its costly grid search. 

It should be noticed that reducing the candidate set size of \Classy{} would have an exponential reduction in its runtimes without much deterioration of its classification performance, as can be seen in Figures~\ref{fig:candidatetime} and \ref{fig:candidateauc}.

\subsection{Discussion}

From the classification and interpretability results of Table~\ref{table:accuracy} and \ref{table:interpretability} it can be seen that \Classy{} is able to provide a good trade-off between classification performance and rule list size. This is particularly the case for multiclass datasets such as \emph{chessbig}, where classical algorithms like JRip find a model with double the number of rules, or in the case of \emph{mushroom}, where CART and C5.0 find more complex models with the same performance. It is interesting to notice that \Classy{} performs better in terms of AUC than accuracy. This shows that when it makes a wrong prediction it does so with a small probability, which is reassuring. Moreover, \Classy{} has only one hyperparameter --- its candidate set  ---  which its tuning is hardly needed as the algorithm has no problem in dealing with large numbers of candidates. This is quite different from the extensive tuning done for the other methods. It is important to observe that \emph{all methods except for \Classy{} were tuned}.

More importantly, in the set of Figures~\ref{fig:candidatesetinfluence}, it is shown that larger candidate sets do not result in worse models, as our formalization in terms of the MDL principle is well-suited to avoid overfitting without the need for cross-validation and/or parameter tuning. In other words, \Classy{} is insensitive to its only parameter --- its candidate set --- making it virtually parameter-free. This is a big advantage, as one can simply run \Classy{} on all training data with as many candidates as possible, without worrying about any parameters. It also means that all training data can be used for training, which is important in case of small data: no data needs to be reserved for validation.  

From Figure~\ref{fig:RelativeCompression} we can observe that better compression corresponds to better classification, which is a strong empirical validation of our formalization. As expected, normalized gain is clearly the best heuristic to use in combination with our greedy rule selection strategy, as it results in better classifiers for $88\%$ of the datasets.

From the runtimes of Figure~\ref{fig:runtime}, it can be seen that \Classy{} runtimes are slower by an order of magnitude than other (fast) algorithms, such as C5.0, CART, and JRip, and similar to SBRL. This is expected for the size of candidate sets used in our experiments, as can be seen in Table~\ref{table:datasets}.

In terms of classification and interpretability, comparing the average ranking with other rule- and tree-based methods in Table~\ref{table:interpretability}, it is shown that \Classy{} performs equally well while also able to find rule lists with less conditions, \emph{without any parameter tunning}. CART creates models with fewer rules that have more conditions per rule, while C5.0 has a high AUC at the expense of over-complex rules. FURIA has a better performance in terms of both standard and balanced accuracy, and worst in terms of AUC, which is expected as it is not a probabilistic classifier. Also, it is hard to compare its interpretability as all its rules can interact with each other, generating a much larger equivalent rule list than \Classy. SBRL on the other hand seems to be able to find simple models that underperform in terms of AUC compared with \Classy, which can be either a result of its formalization or because it cannot use larger candidate sets. The experiments also revealed that the Poisson distribution used as prior in SBRL, for the number of conditions per rule and the number of rules, creates tight constraints from which the results hardly deviate. Our results suggest that if the `optimal' values for these hyperparameters are not known in advance, the best model may not be found. An indicative example of this is the \emph{tictactoe} dataset in Table \ref{table:accuracy}, a deterministic dataset for which SBRL can only find the right amount of rules and logical conditions per rule when given these exact values in advance. The results obtained with \Classy{} demonstrate that using the universal prior for integers alleviates this strong dependence on hyperparameter tuning.

In terms of overfitting, Table~\ref{table:overfitting} shows that \Classy{} has a tendency to select models that generalize well and that are not overconfident in the training set. It obtains low differences between training and test compared with the other rule- and tree-based models.

\section{Conclusions and future work}
\label{sec:conclusion}

We proposed a novel formalization of the multiclass classification problem using probabilistic rule lists and the minimum description length (MDL) principle. Our problem formulation allows for parameter-free model selection and naturally trades off model complexity with predictive accuracy, effectively avoiding overfitting. To find solutions to this problem, we introduced the heuristic \Classy{} algorithm, which greedily constructs rule lists using the MDL-based criterion.

We empirically demonstrated, on a variety of datasets, that \Classy{} finds probabilistic rule lists that perform on par with state-of-the-art interpretable classifiers with respect to predictive accuracy, despite the fact that some form of hyperparameter tuning is done for all methods except for \Classy{}. Moreover, the models found by our approach were shown to be more compact than those obtained by the other methods, which is expected to make them more understandable in practice. Finally, compression was shown to strongly correlate with predictive accuracy, which can be regarded as an empirical validation of the MDL-based selection criterion.

Directions for future work include, for instance, the following:
\begin{itemize}
\item Bridge the gap between both kinds of search methods used to learn rule lists from data in a principled way, namely optimal strategies --- accurate but slow --- and greedy methods --- fast but imperfect.
\item Extend our MDL formulation to other types of data and/or tasks, such as continuous data and regression problems.
\end{itemize}

\section*{Acknowledgment}
This work is part of the research programme Indo-Dutch Joint Research Programme for ICT 2014 with project number 629.002.201, SAPPAO, which is (partly) financed by the Netherlands Organisation for Scientific Research (NWO).

\bibliographystyle{apa}
\bibliography{bibtest}

\begin{thebibliography}{}

\bibitem[\protect\astroncite{Agrawal et~al.}{1993}]{agrawal1993mining}
Agrawal, R., Imieli{\'n}ski, T., and Swami, A. (1993).
\newblock Mining association rules between sets of items in large databases.
\newblock In {\em Acm sigmod record}, volume~22, pages 207--216. ACM.

\bibitem[\protect\astroncite{Alcala-Fdez et~al.}{2011}]{alcala2011fuzzy}
Alcala-Fdez, J., Alcala, R., and Herrera, F. (2011).
\newblock A fuzzy association rule-based classification model for
  high-dimensional problems with genetic rule selection and lateral tuning.
\newblock {\em IEEE Transactions on Fuzzy systems}, 19(5):857--872.

\bibitem[\protect\astroncite{Angelino et~al.}{2017}]{angelino2017learning}
Angelino, E., Larus-Stone, N., Alabi, D., Seltzer, M., and Rudin, C. (2017).
\newblock Learning certifiably optimal rule lists.
\newblock In {\em KDD'17}. ACM.

\bibitem[\protect\astroncite{Aoga et~al.}{2018}]{aogafinding}
Aoga, J. O.~R., Guns, T., Nijssen, S., and Schaus, P. (2018).
\newblock Finding probabilistic rule lists using the minimum description length
  principle.
\newblock In {\em DS'18}.

\bibitem[\protect\astroncite{Bellodi and Riguzzi}{2015}]{bellodi2015structure}
Bellodi, E. and Riguzzi, F. (2015).
\newblock Structure learning of probabilistic logic programs by searching the
  clause space.
\newblock {\em Theory and Practice of Logic Programming}, 15(2):169--212.

\bibitem[\protect\astroncite{Borgelt}{2003}]{borgelt2003efficient}
Borgelt, C. (2003).
\newblock Efficient implementations of apriori and eclat.
\newblock In {\em FIMI’03: Proceedings of the IEEE ICDM workshop on frequent
  itemset mining implementations}.

\bibitem[\protect\astroncite{Breiman et~al.}{1984}]{breiman1984classification}
Breiman, L., Friedman, J., Stone, C.~J., and Olshen, R.~A. (1984).
\newblock {\em Classification and regression trees}.
\newblock CRC press.

\bibitem[\protect\astroncite{Brodersen et~al.}{2010}]{brodersen2010balanced}
Brodersen, K.~H., Ong, C.~S., Stephan, K.~E., and Buhmann, J.~M. (2010).
\newblock The balanced accuracy and its posterior distribution.
\newblock In {\em 2010 20th International Conference on Pattern Recognition},
  pages 3121--3124. IEEE.

\bibitem[\protect\astroncite{Budhathoki and
  Vreeken}{2015}]{budhathoki2015difference}
Budhathoki, K. and Vreeken, J. (2015).
\newblock The difference and the norm -- characterising similarities and
  differences between databases.
\newblock In {\em ECMLPKDD'15}, pages 206--223. Springer.

\bibitem[\protect\astroncite{Cohen}{1995}]{cohen1995fast}
Cohen, W.~W. (1995).
\newblock Fast effective rule induction.
\newblock In {\em Machine Learning Proceedings 1995}, pages 115--123. Elsevier.

\bibitem[\protect\astroncite{Dem{\v{s}}ar}{2006}]{demvsar2006statistical}
Dem{\v{s}}ar, J. (2006).
\newblock Statistical comparisons of classifiers over multiple data sets.
\newblock {\em Journal of Machine learning research}, 7(Jan):1--30.

\bibitem[\protect\astroncite{Doshi-Velez and Kim}{2017}]{doshi17interpretable}
Doshi-Velez, F. and Kim, B. (2017).
\newblock Towards a rigorous science of interpretable machine learning.
\newblock {\em arXiv:1702.08608 [stat.ML]}.

\bibitem[\protect\astroncite{Fernandez et~al.}{2015}]{fernandez2015revisiting}
Fernandez, A., Lopez, V., del Jesus, M.~J., and Herrera, F. (2015).
\newblock Revisiting evolutionary fuzzy systems: Taxonomy, applications, new
  trends and challenges.
\newblock {\em Knowledge-Based Systems}, 80:109--121.

\bibitem[\protect\astroncite{Friedman}{1937}]{friedman1937use}
Friedman, M. (1937).
\newblock The use of ranks to avoid the assumption of normality implicit in the
  analysis of variance.
\newblock {\em Journal of the american statistical association},
  32(200):675--701.

\bibitem[\protect\astroncite{F{\"u}rnkranz
  et~al.}{2012}]{furnkranz2012foundations}
F{\"u}rnkranz, J., Gamberger, D., and Lavra{\v{c}}, N. (2012).
\newblock {\em Foundations of rule learning}.
\newblock Springer Science \& Business Media.

\bibitem[\protect\astroncite{Garc{\'\i}a-Borroto
  et~al.}{2014}]{garcia2014survey}
Garc{\'\i}a-Borroto, M., Mart{\'\i}nez-Trinidad, J.~F., and Carrasco-Ochoa,
  J.~A. (2014).
\newblock A survey of emerging patterns for supervised classification.
\newblock {\em Artificial Intelligence Review}, 42(4):705--721.

\bibitem[\protect\astroncite{Gelman et~al.}{2013}]{gelman2013bayesian}
Gelman, A., Stern, H.~S., Carlin, J.~B., Dunson, D.~B., Vehtari, A., and Rubin,
  D.~B. (2013).
\newblock {\em Bayesian data analysis}.
\newblock Chapman and Hall/CRC.

\bibitem[\protect\astroncite{Gr{\"u}nwald}{2007}]{grunwald2007minimum}
Gr{\"u}nwald, P.~D. (2007).
\newblock {\em The minimum description length principle}.
\newblock MIT press.

\bibitem[\protect\astroncite{Holm}{1979}]{holm1979simple}
Holm, S. (1979).
\newblock A simple sequentially rejective multiple test procedure.
\newblock {\em Scandinavian journal of statistics}, pages 65--70.

\bibitem[\protect\astroncite{H{\"u}hn and
  H{\"u}llermeier}{2009}]{huhn2009furia}
H{\"u}hn, J. and H{\"u}llermeier, E. (2009).
\newblock Furia: an algorithm for unordered fuzzy rule induction.
\newblock {\em Data Mining and Knowledge Discovery}, 19(3):293--319.

\bibitem[\protect\astroncite{Huysmans et~al.}{2011}]{huysmans2011empirical}
Huysmans, J., Dejaeger, K., Mues, C., Vanthienen, J., and Baesens, B. (2011).
\newblock An empirical evaluation of the comprehensibility of decision table,
  tree and rule based predictive models.
\newblock {\em Decision Support Systems}, 51(1):141--154.

\bibitem[\protect\astroncite{Iman and Davenport}{1980}]{iman1980approximations}
Iman, R.~L. and Davenport, J.~M. (1980).
\newblock Approximations of the critical region of the fbietkan statistic.
\newblock {\em Communications in Statistics-Theory and Methods}, 9(6):571--595.

\bibitem[\protect\astroncite{Jim{\'e}nez et~al.}{2014}]{jimenez2014multi}
Jim{\'e}nez, F., S{\'a}nchez, G., and Ju{\'a}rez, J.~M. (2014).
\newblock Multi-objective evolutionary algorithms for fuzzy classification in
  survival prediction.
\newblock {\em Artificial intelligence in medicine}, 60(3):197--219.

\bibitem[\protect\astroncite{{Kralj Novak} et~al.}{2009}]{novak09sd}
{Kralj Novak}, P., Lavra\v{c}, N., and Webb, G. (2009).
\newblock Supervised descriptive rule discovery: A unifying survey of contrast
  set, emerging pattern and subgroup mining.
\newblock {\em Journal of Machine Learning Research}, 10:377--403.

\bibitem[\protect\astroncite{Lakkaraju
  et~al.}{2016}]{lakkaraju2016interpretable}
Lakkaraju, H., Bach, S.~H., and Leskovec, J. (2016).
\newblock Interpretable decision sets: A joint framework for description and
  prediction.
\newblock In {\em KDD'16}. ACM.

\bibitem[\protect\astroncite{Lakkaraju and Rudin}{2016}]{lakkarajulearning}
Lakkaraju, H. and Rudin, C. (2016).
\newblock Learning cost-effective and interpretable treatment regimes for
  judicial bail decisions.
\newblock In {\em NIPS 2016}.

\bibitem[\protect\astroncite{Lakkaraju and Rudin}{2017}]{lakkaraju2017learning}
Lakkaraju, H. and Rudin, C. (2017).
\newblock Learning cost-effective and interpretable treatment regimes.
\newblock In {\em Artificial Intelligence and Statistics}.

\bibitem[\protect\astroncite{Letham et~al.}{2015}]{letham2015interpretable}
Letham, B., Rudin, C., McCormick, T.~H., Madigan, D., et~al. (2015).
\newblock Interpretable classifiers using rules and bayesian analysis: Building
  a better stroke prediction model.
\newblock {\em The Annals of Applied Statistics}, 9(3):1350--1371.

\bibitem[\protect\astroncite{Li et~al.}{2001}]{li2001cmar}
Li, W., Han, J., and Pei, J. (2001).
\newblock Cmar: Accurate and efficient classification based on multiple
  class-association rules.
\newblock In {\em Data Mining, 2001. ICDM 2001, Proceedings IEEE International
  Conference on}, pages 369--376. IEEE.

\bibitem[\protect\astroncite{Lou et~al.}{2012}]{lou2012intelligible}
Lou, Y., Caruana, R., and Gehrke, J. (2012).
\newblock Intelligible models for classification and regression.
\newblock In {\em KDD'12}, pages 150--158. ACM.

\bibitem[\protect\astroncite{Ma and Liu}{1998}]{ma1998integrating}
Ma, B. L. W. H.~Y. and Liu, B. (1998).
\newblock Integrating classification and association rule mining.
\newblock In {\em KDD'98}.

\bibitem[\protect\astroncite{Molnar}{2018}]{molnar2018interpretable}
Molnar, C. (2018).
\newblock Interpretable machine learning.
\newblock {\em A Guide for Making Black Box Models Explainable}.

\bibitem[\protect\astroncite{Polaka et~al.}{2017}]{polaka2017constructing}
Polaka, I., Ga{\v{s}}enko, E., Barash, O., Haick, H., and Leja, M. (2017).
\newblock Constructing interpretable classifiers to diagnose gastric cancer
  based on breath tests.
\newblock {\em Procedia Computer Science}, 104.

\bibitem[\protect\astroncite{Provost and Domingos}{2000}]{provost2000well}
Provost, F. and Domingos, P. (2000).
\newblock Well-trained pets: Improving probability estimation trees.

\bibitem[\protect\astroncite{Quinlan}{2014}]{quinlan2014c4}
Quinlan, J.~R. (2014).
\newblock {\em C4. 5: programs for machine learning}.
\newblock Elsevier.

\bibitem[\protect\astroncite{Ribeiro et~al.}{2016}]{ribeiro2016should}
Ribeiro, M.~T., Singh, S., and Guestrin, C. (2016).
\newblock Why should i trust you?: Explaining the predictions of any
  classifier.
\newblock In {\em KDD'16}, pages 1135--1144. ACM.

\bibitem[\protect\astroncite{Ribeiro et~al.}{2018}]{ribeiro2018anchors}
Ribeiro, M.~T., Singh, S., and Guestrin, C. (2018).
\newblock Anchors: High-precision model-agnostic explanations.
\newblock In {\em Proceedings of the Thirty-Second AAAI Conference on
  Artificial Intelligence (AAAI)}.

\bibitem[\protect\astroncite{Rissanen}{1978}]{rissanen78}
Rissanen, J. (1978).
\newblock Modeling by shortest data description.
\newblock {\em Automatica}, 14(5).

\bibitem[\protect\astroncite{Rissanen}{1983}]{rissanen1983universal}
Rissanen, J. (1983).
\newblock A universal prior for integers and estimation by minimum description
  length.
\newblock {\em The Annals of statistics}, pages 416--431.

\bibitem[\protect\astroncite{van Leeuwen and
  Galbrun}{2015}]{van2015association}
van Leeuwen, M. and Galbrun, E. (2015).
\newblock Association discovery in two-view data.
\newblock {\em IEEE Transactions on Knowledge and Data Engineering}, 27(12).

\bibitem[\protect\astroncite{van Leeuwen and Vreeken}{2014}]{van2014mining}
van Leeuwen, M. and Vreeken, J. (2014).
\newblock Mining and using sets of patterns through compression.
\newblock In {\em Frequent Pattern Mining}, pages 165--198. Springer.

\bibitem[\protect\astroncite{Vreeken et~al.}{2011}]{vreeken2011krimp}
Vreeken, J., van Leeuwen, M., and Siebes, A. (2011).
\newblock Krimp: mining itemsets that compress.
\newblock {\em Data Mining and Knowledge Discovery}, 23(1):169--214.

\bibitem[\protect\astroncite{Wang and Karypis}{2005}]{wang2005harmony}
Wang, J. and Karypis, G. (2005).
\newblock Harmony: Efficiently mining the best rules for classification.
\newblock In {\em Proceedings of the 2005 SIAM International Conference on Data
  Mining}, pages 205--216. SIAM.

\bibitem[\protect\astroncite{Wang et~al.}{2016}]{wang2016bayesian}
Wang, T., Rudin, C., Velez-Doshi, F., Liu, Y., Klampfl, E., and MacNeille, P.
  (2016).
\newblock Bayesian rule sets for interpretable classification.
\newblock In {\em Data Mining (ICDM), 2016 IEEE 16th International Conference
  on}, pages 1269--1274. IEEE.

\bibitem[\protect\astroncite{Webb}{2007}]{webb07sigpatts}
Webb, G.~I. (2007).
\newblock Discovering significant patterns.
\newblock {\em Machine Learning}, 68(1):1--33.

\bibitem[\protect\astroncite{Yang et~al.}{2017}]{yang2017scalable}
Yang, H., Rudin, C., and Seltzer, M. (2017).
\newblock Scalable bayesian rule lists.
\newblock In {\em Proceedings of the 34th International Conference on Machine
  Learning-Volume 70}, pages 3921--3930. JMLR. org.

\bibitem[\protect\astroncite{Zeng et~al.}{2017}]{zeng2017interpretable}
Zeng, J., Ustun, B., and Rudin, C. (2017).
\newblock Interpretable classification models for recidivism prediction.
\newblock {\em Journal of the Royal Statistical Society: Series A (Statistics
  in Society)}, 180(3).

\bibitem[\protect\astroncite{Zhang et~al.}{2000}]{zhang2000information}
Zhang, X., Dong, G., and Ramamohanarao, K. (2000).
\newblock Information-based classification by aggregating emerging patterns.
\newblock In {\em IDEAL}, pages 48--53. Springer.

\bibitem[\protect\astroncite{Zimmermann and
  Nijssen}{2014}]{zimmermann14supervised}
Zimmermann, A. and Nijssen, S. (2014).
\newblock Supervised pattern mining and applications to classification.
\newblock In {\em Frequent Pattern Mining}. Springer.

\end{thebibliography}

\end{document}